\documentclass[]{elsarticle}
\usepackage{lineno, hyperref}
\usepackage{multirow}
\usepackage{amsmath}

\usepackage{tikz}
\usetikzlibrary{trees}

\usepackage[most]{tcolorbox}

\usepackage{caption}
\usepackage{subcaption}

\usepackage{rotating}

\usepackage{xcolor}

\usepackage[ruled, linesnumbered]{algorithm2e}

\biboptions{authoryear}

\journal{Expert Systems with Applications}
\date{May 28, 2022}

\begin{document}

\begin{frontmatter}
\title{Multi-Resolution, Multi-Horizon Distributed Solar PV Power Forecasting with Forecast Combinations}

% \author{Author(s)}
\author{Maneesha Perera\corref{cor1}} 
\ead{maneesha.perera@student.unimelb.edu.au}
\address{Department of Mechanical Engineering, School of Electrical, Mechanical and Infrastructure Engineering, The University of Melbourne, Melbourne, Australia}

\author{Julian De Hoog}
\ead{julian.dehoog@unimelb.edu.au}
\address{Department of Mechanical Engineering, School of Electrical, Mechanical and Infrastructure Engineering, The University of Melbourne, Melbourne, Australia}

\author{Kasun Bandara}
\ead{kasun.bandara@unimelb.edu.au}
\address{School of Computing and Information Systems, Melbourne Centre for Data Science, The University of Melbourne, Melbourne, Australia}

\author{Saman Halgamuge}
\ead{saman@unimelb.edu.au}
\address{Department of Mechanical Engineering, School of Electrical, Mechanical and Infrastructure Engineering, The University of Melbourne, Melbourne, Australia}

\cortext[cor1]{Corresponding author}

\nonumnote{This manuscript was accepted for publication on May 28, 2022. This is the accepted version of the manuscript. The published version of the manuscript is available at: \url{https://doi.org/10.1016/j.eswa.2022.117690}}
\nonumnote{© 2022. This manuscript version is made available under the CC-BY-NC-ND 4.0 license \url{https://creativecommons.org/licenses/by-nc-nd/4.0/}}

\newpageafter{author}
%%%%% ABSTRACT %%%%%
\begin{abstract}

Distributed, small-scale solar photovoltaic (PV) systems are being installed at a rapidly increasing rate. This can cause major impacts on distribution networks and energy markets. As a result, there is a significant need for improved forecasting of the power generation of these systems at different time resolutions and horizons. However, the performance of forecasting models depends on the resolution and horizon. Forecast combinations (ensembles), that combine the forecasts of multiple models into a single forecast may be robust in such cases. Therefore, in this paper, we provide comparisons and insights into the performance of five state-of-the-art forecast models and existing forecast combinations at multiple resolutions and horizons. We propose a forecast combination approach based on particle swarm optimization (PSO) that will enable a forecaster to produce accurate forecasts for the task at hand by weighting the forecasts produced by individual models. Furthermore, we compare the performance of the proposed combination approach with existing forecast combination approaches. A comprehensive evaluation is conducted using a real-world residential PV power data set measured at 25 houses located in three locations in the United States. The results across four different resolutions and four different horizons show that the PSO-based forecast combination approach outperforms the use of any individual forecast model and other forecast combination counterparts, with an average Mean Absolute Scaled Error reduction by 3.81\% compared to the best performing individual model. Our approach enables a solar forecaster to produce accurate forecasts for their application regardless of the forecast resolution or horizon.

\end{abstract}
%%%%% KEYWORDS %%%%% 
\begin{keyword}
Solar Photovoltaic Power Forecasting, Forecast Combinations, Ensembles, Particle Swarm Optimization, Multi-Resolution, Multi-Horizon
\end{keyword}

\end{frontmatter}

%%%%% INTRODUCTION %%%%% 
\section{Introduction}\label{sec:introduction}

Solar photovoltaic (PV) generation is the fastest growing form of energy generation today~\citep{iea2019}. According to the International Energy Agency, renewable energy capacity is projected to increase by 1200GW between 2019 and 2024. Solar PV is projected to account for 60\% of this growth, and much of this will be due to the emergence of small-scale, distributed systems on the rooftops of homes and businesses~\citep{iea2019}. The increasing uptake of these solar PV systems brings with it many challenges and opportunities. Distribution network operators are finding it difficult to accommodate the voltage fluctuations caused by large amounts of solar PV~\citep{arena}. Distributed solar PV is starting to have an impact on the net demand at wholesale market level, leading to market operators having to increasingly take distributed solar PV into account in their forecasting models~\citep{aemo}. Increasingly, solar PV is paired with energy storage to help accommodate variability of solar resources, but energy storage systems need to be managed carefully to maximise their economic return~\citep{Abdulla2018OptimalDegradation}.

For many applications there is a strong need for accurate forecasts of solar PV power generation. However, the \textit{forecast horizon} and the \textit{forecast resolution} may change significantly from one application to another. Day ahead market trading systems, for example, may require a day ahead, hourly solar PV power forecast. Energy storage optimisation systems, on the other hand, may benefit from forecasts that are only an hour ahead, at a 5-minute resolution. Many additional forecasting applications exist; Table~\ref{tab:applications} provides some further examples.

\begin{table}[h]
\centering
\small
\begin{tabular}{|l|l|l|l|}
\hline
\textbf{Horizon} & \textbf{Forecast Span} & \textbf{Resolution} & \textbf{Application}                                                                                \\ \hline
Intra hour          & 1-60 minutes  & 1, 5, 15, 30 minutes &\begin{tabular}[c]{@{}l@{}}Real time control \\ Ramp rate control\\ Variability management\end{tabular} \\ \hline
Intra day           & 1-6 hours &  Hourly & \begin{tabular}[c]{@{}l@{}}Load following \\ Demand response scheduling \end{tabular}                                                                                           \\\hline
Day ahead           & 1-7 days    & 1, 3 days & \begin{tabular}[c]{@{}l@{}}Transmission scheduling\\ Economic dispatch\end{tabular}      \\\hline               
\end{tabular}
\caption{Forecasting requirements for some applications, having varying temporal resolutions and horizons.}
\label{tab:applications}
\end{table}

\begin{figure}[t]
    \centering
    \includegraphics[width=0.9\textwidth]{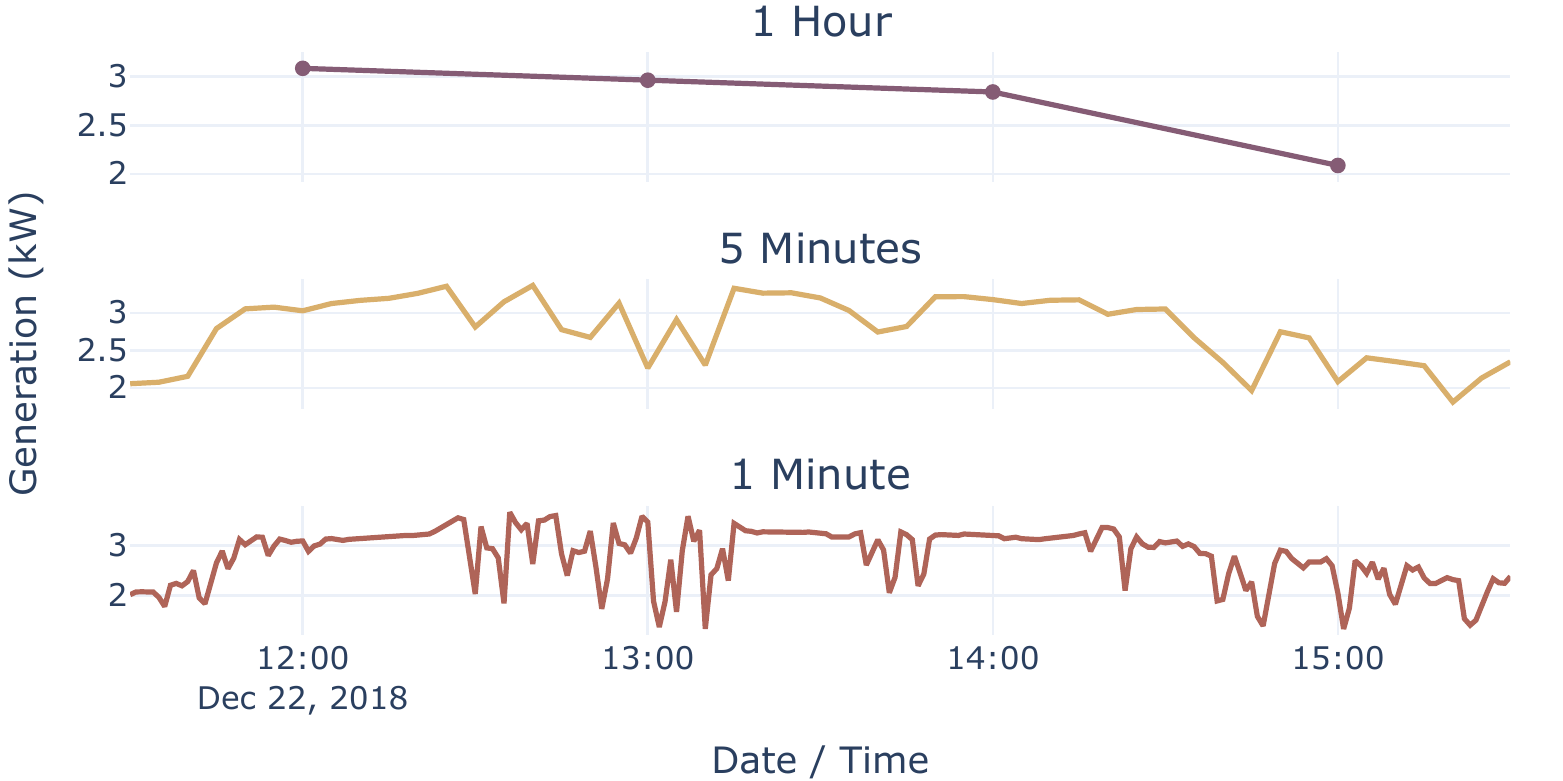}
    \caption{An example of PV power data at different time resolutions of a PV system in a house located in Austin in the United States. Sudden fluctuations in the power generation can be observed at high resolutions (1 minute, 5 minutes), while at low resolutions (1 hour) some of these patterns are no longer apparent.}
    \label{fig:resolutions}
\end{figure}

 The temporal resolution and horizon of the data has a direct impact on the forecast performance. Figure~\ref{fig:resolutions} shows an example of residential PV power data captured at three different time resolutions. As seen in the Figure, sudden fluctuations in the power generation can be observed at high resolutions, while at low resolutions some of these patterns are no longer visible. Many solar power forecasting studies focus on improving the forecast accuracy for one specific horizon, or for several horizons but using one specific resolution~\citep{Antonanzas2016ReviewForecasting, Long2014AnalysisApproaches, DeGiorgi2016ComparisonMachine}. In this work, we aim to broaden the understanding of how different types of forecasting models perform across different resolutions and horizons, and provide insights that will be of practical value to stakeholders in the energy sector.

Many approaches exist for solar power forecasting. These include numerical weather prediction (NWP) models \citep{Mathiesen2011EvaluationStates, Mayer2021ExtensiveForecasting}, remote sensing methods (e.g., satellite imagery) \citep{Si2021PhotovoltaicPosition}, statistical methods \citep{Pedregal2021AdjustedIrradiance}, machine learning \citep{Yagli2019AutomaticModels}, deep learning methods \citep{duPlessis2021Short-termBehaviour, Korkmaz2021SolarNet:Forecasting} and hybrid approaches \citep{Qu2021AForecasting, Lan2019Day-aheadNetwork}. Data-driven approaches (e.g., statistical methods, machine learning methods) are the most commonly used techniques in PV power forecasting literature \citep{Mayer2021ExtensiveForecasting}. In most cases, forecast combinations (also referred to as ensembles, or hybrid approaches), that combine the forecasts of multiple models into a single forecast, have been shown to improve the forecast performance~\citep{ Ren2015EnsembleReview, Sperati2016AnForecasting}. There exist different ways to create forecast combinations using data-driven approaches. One such way is to generate forecasts using multiple models (often referred to as base forecasters) and combine these forecasts by averaging, or by using a weighted aggregation. In a weighted aggregation, finding the optimal weights to combine the forecasts produced by the base forecasters plays a major role in determining the forecast accuracy. Previous works on solar PV power forecasting have considered these weights to be a convex combination, i.e., weight values are non-negative, between 0 and 1, and they sum to 1~\citep{Atiya2020WhyWell, Liu2019ASystems}.

While there exists various different forecasting models to forecast the solar power, these models have a varying performance at different temporal resolutions and horizons~\citep{Antonanzas2016ReviewForecasting, Reikard2009PredictingForecasts}. This creates a challenge for a forecaster in determining which forecasting model to be used for the task at hand. Therefore, in this paper, we introduce a forecast combination approach based on Particle Swarm Optimization (PSO) that finds a set of optimal weights to combine the forecasts produced by diverse state-of-the-art forecasting models. Our approach will enable a forecaster to produce accurate solar PV power forecasts regardless of the resolution and the horizon of the task of interest. We further explore and discuss how the conditions imposed when finding these weights may impact the forecasting performance. We conduct a comprehensive evaluation of the forecast accuracy of (i) base forecasters, (ii) forecast combinations with weights across four different resolutions (1 minute, 5 minutes, 1 hour, 1 day) and across four different horizons (5 minutes, 1 hour, 1 day and 3 days) using a real residential solar PV power data measured at 25 houses located in three different locations in the United States. All code used to develop the models and generate the results presented in this paper is made available\footnote{\label{code}\url{https://github.com/ManeeshaPerera/solar-forecasting-framework}}.

The rest of the paper is structured as follows. Section~\ref{sec:relatedwork} describes related work on forecast combinations in PV power forecasting and temporal resolutions, horizons studied in the literature. Section~\ref{sec:experiments} describes the experimental data used in this study. Section~\ref{sec:method} explains the forecast combination approach and Section~\ref{sec:results} discusses the results. Finally, Section~\ref{sec:conclusion} presents our conclusions.

%%%%% RELATED WORK %%%%% 
\section{Related Work}\label{sec:relatedwork}

Solar forecasting studies have progressed rapidly in recent years \citep{Antonanzas2016ReviewForecasting, Yang2018HistoryMining, Ahmed2020AOptimization}. In the following we discuss studies relevant to the context of this paper: forecast resolutions and horizons in solar forecasting studies and forecast combinations with data-driven techniques.

\subsection{Forecast Resolutions and Horizons}\label{sec:resolutions}

\begin{figure}[th]
    \centering
    \includegraphics[width=\textwidth]{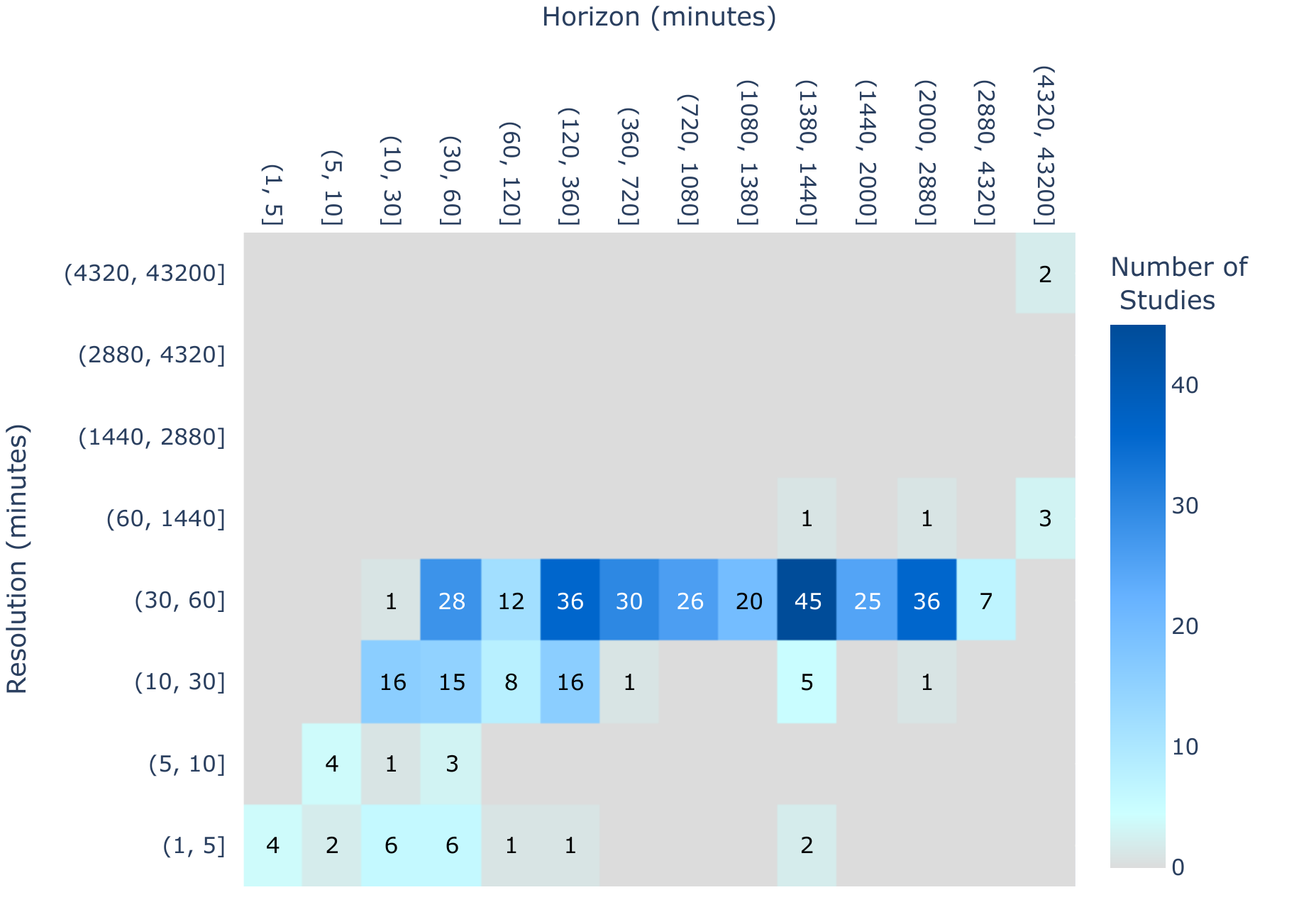}
    \caption{Number of studies conducted at different resolutions and horizons from 104 studies. Studies that have considered multiple resolutions/ horizons are represented in multiple cells corresponding to the respective resolution and horizon pairs studied. There are 10 unique resolutions and 84 unique horizons considering all the studies. For readability of the figure, the resolution and horizons are provided in intervals (e.g., (5, 10] resolution indicates work that has studied a resolution of 5 (exclusive) to 10 (inclusive) minutes).}
    \label{fig:heatmap}
\end{figure}

We analysed the horizons and resolutions of 104 studies in solar forecasting, to understand which of these are most common and have the greatest number of applications. Figure~\ref{fig:heatmap} presents an annotated heat map showing the outcome of this analysis. While \cite{Antonanzas2016ReviewForecasting} (presented in Table 2 in their paper) previously reviewed many of these studies from 2007 to 2016, we have further included several recent works presented in Table \ref{tab:papers}. A complete list of all 104 studies considered in this analysis is provided here \footnote{\url{https://github.com/ManeeshaPerera/solar-forecasting-framework/blob/master/resolution_and_horizons_papers.pdf}}. We note that Figure \ref{fig:heatmap} is only a representation of a subset of solar forecasting studies and does not include all solar forecasting studies in the literature. However, this analysis provides some insight as to the most studied resolutions and horizons in the literature. As can be seen, most work is conducted at an hourly resolution. This is likely due to both (i) many applications requiring this level of resolution, and (ii) unavailability of data at higher resolutions in many situations. The most studied resolution and horizon pair is forecasting a horizon of 1440 minutes (1 day) at a 60-minute resolution. This is likely due to the fact that many energy markets require forecasts having this resolution and horizon as part of their day-ahead market operations~\citep{Antonanzas2017TheMarket, BrancucciMartinez-Anido2016TheImprovement}. 

\begin{table}[!th]
\small
\begin{tabular}{|l|l|l|}
\hline
\textbf{Author}                              & \textbf{Resolution} & \textbf{Horizon}                                                                 \\ \hline
\cite{Abdel-Nasser2017AccurateLSTM-RNN}   & 1 hour              & 1 hour                                                                           \\
\cite{Nespoli2019Day-aheadTechniques}              & 15 minutes          & 1 day                                                                            \\
\cite{Lee2019RecurrentInformation}                 & 1 hour              & 14 hours                                                                         \\
\cite{Liu2019ASystems}                           & 5 minutes           & 1 day                                                                            \\
\cite{VanDeventer2019}                           & 1 hour              & 1 hour                                                                           \\
\cite{Lan2019Day-aheadNetwork}                     & 1 hour              & 1 day                                                                            \\
\cite{Zang2020Day-aheadLearning}                   & 1 hour              & 1 day                                                                            \\
\cite{Heo2021Multi-channelForecasting}              & 30 days             & 30 days                                                                          \\
\cite{Korkmaz2021SolarNet:Forecasting}              & 1 hour              & 1 hour, 2 hours, 3 hours                                                           \\
\cite{Qu2021AForecasting}                          & 5 minutes           & 1 day                                                                            \\
\cite{Lai2021AForecasting}                        & 1 hour              & 1 hour                                                                           \\
\cite{DeHoog2021CharacteristicProfile}             & 15 minutes          & 1 day                                                                            \\
\cite{duPlessis2021Short-termBehaviour}           & 15 minutes          & 1 hour, 2 hours, 3 hours,\\
& & 4 hours, 5 hours, 6 hours \\
\cite{Kumari2021LongForecasting}                   & 1 hour              & 1 hour                                                                           \\
\cite{Mayer2021ExtensiveForecasting}                & 15 minutes          & 1 day, 2 days                                                                     \\
\multirow{2}{*}{\cite{Cannizzaro2021SolarLearning}} & 15 minutes          & 15, 30, 40, 45 minutes, \\
& & 1, 3, 12 hours, 1 day \\
                                             & 1 hour              & 1, 3, 12, hours, 1 day                                                           \\
\cite{Castangia2021AForecasting}                   & 15 minutes          & 15, 30, 45, 60, 75, 90, 105, \\
& & 120, 135, 150, 165, 180, \\
& & 195, 210, 225, 240 minutes \\
\cite{Acikgoz2022AForecasting}                     & 1 hour              & 1 hour, 2 hours, 3 hours                                                           \\
\cite{Aicardi2022AIrradiation}                      & 1 hour              & 1 hour, 2 hours, 3 hours, 4 hours, \\
& &  5 hours \\
\hline                                         
\end{tabular}
\caption{Resolution and horizon of some recent solar forecasting studies in literature.}
\label{tab:papers}
\end{table}

\subsection{Forecast Combinations}\label{sec:forecastingtechniques}

The choice of which forecasting method to use is highly dependent on the resolution of the data and the required forecast horizon~\citep{Alskaif2020AEstimation, Das2018ForecastingReview}. To this end, different types of forecasting methods have been studied in the literature. Hybrid approaches (also known as ensembles, or forecast combinations) that combine forecasts produced by multiple methods have shown promising results compared to other approaches~\citep{Das2018ForecastingReview, Antonanzas2016ReviewForecasting}. The intuition behind hybrid modelling is that different forecasting methods are suitable to model distinctive attributes of the data and therefore combining them will result in an improved forecast.

Hybrid approaches can be created in various ways~\citep{Ren2015EnsembleReview}. One such way is to decompose the forecasting task into sub-tasks and solve these sub-tasks independently, either using the same model or using different models. The final forecasts are derived by appending the forecasts of all sub-tasks. Time series decomposition techniques like wavelet decomposition have been used in this way~\citep{Mandal2012ForecastingTechniques, Haque2013SolarApproach, Alhakeem2015AIntervals}. Hybrid approaches have also been created by separately modelling the linear and non-linear components of the PV power data~\citep{Bouzerdoum2013APlant}.  

Another (more common) way is to conduct the forecasting task repeatedly using a set of diverse models often known as base forecasters (these may also be called base learners, or stand-alone models, depending on the study). The final forecasts are derived by averaging the base forecasters, or by using a weighted aggregation of these. Having diversity in the base forecasters is typically important in this kind of approach. Such diversity can be achieved in various ways. For instance,~\cite{Rana2016UnivariateForecasting} and ~\cite{Raza2018AnGrids} achieved the diversity in the ensemble approaches by implementing an ensemble of neural networks (NN) with different parameters and architectures to forecast the PV power generation. Optimization methods like evolutionary algorithms have been used in such cases to achieve diversity by creating models with varying parameters/ architectures~\citep{VanDeventer2019, Raza2016OnForecast}. \cite{VanDeventer2019} employed a genetic algorithm to find the optimal parameters for multiple support vector machine (SVM) classifiers to forecast the PV power. 

The aggregation approach (averaging, or weighted aggregation) also plays a significant role when creating forecast combinations. The most simple and widely used combination approach in the literature is to take the average of the forecasts produced by all base forecasters~\citep{Genre2013CombiningAverage}. However, studies have also looked into finding optimal weights to combine these forecasts. A study conducted by \cite{Liu2019ASystems} proposed a recursive ensemble strategy to combine three forecast models.
In their work, weights were found in an iterative approach where the forecasts of the worst model (selected based on the error) is replaced by the average prediction of all other base forecasters in every iteration. This recursive approach was continued until a defined threshold value was met and the weight combination with the minimum error is selected. Another study was conducted by \cite{Galicia2019Multi-stepLearning} in which they applied a weighted least square method to find the optimal weights to combine the forecasts produced by three base forecasters.

In these types of studies, the set of optimal weights are almost always considered to be a convex combination, meaning that the weights are non-negative, are all less than 1, and sum up to 1. The impact of these weight conditions on the forecast accuracy has been studied with other time series data (e.g. economic time series)~\citep{Genre2013CombiningAverage}. For instance, in a study conducted by \cite{Genre2013CombiningAverage} not imposing convexity conditions on weights showed a better performance for inflation rate forecasts. However, such conditions on the weights have not been extensively evaluated in the context of solar PV power forecasting, which is different from other time series data in terms of seasonality and variability. In this work, we experimentally study the impact of such weight conditions in forecast combinations for solar power forecasting.

\section{Evaluation Framework}\label{sec:experiments}

In this section, we describe how we set up our evaluation framework: the data used, the choices of resolutions and horizons to be compared, and the error metrics applied.

\subsection{Data}

The solar PV power dataset used in this study is from the Pecan Street's Dataport~\citep{pecanStreet}. Dataport contains energy data collected at a 1-minute resolution for a total of 73 households in the United States. These households are located in Austin, New York and California. A subset of these houses (38 houses) contains PV power generation data. Data for 25 houses were selected from this subset after removing houses that contained missing values for more than 3 consecutive days or greater than 0.5\% of the total data points in the time series. The selected data was then pre-processed by interpolating any existing missing values and removing negative power generation values. A summary of the extracted data is presented in Table~\ref{tab:solarData}.

\begin{table}[th]
\centering
\begin{tabular}{|c|c|c|c|}
\hline
\multicolumn{1}{|l|}{\multirow{2}{*}{\textbf{Location}}} & \multicolumn{1}{l|}{\multirow{2}{*}{\textbf{\begin{tabular}[c]{@{}c@{}}Number of\\ Houses\end{tabular}}}} & \multicolumn{2}{c|}{\textbf{\begin{tabular}[c]{@{}c@{}}Data Characteristics \\ per House\end{tabular}}}                                         \\ \cline{3-4} 
\multicolumn{1}{|l|}{}                                   & \multicolumn{1}{l|}{}                                                                                     & \textbf{Duration}                                                               & \textbf{\begin{tabular}[c]{@{}c@{}}Time\\  Points\end{tabular}} \\ \hline
Austin                                                   & 10                                                                                                        & \begin{tabular}[c]{@{}c@{}}1 Year\\ January 2018 - December 2018\end{tabular} & 525,600                                                         \\
New York                                                 & 14                                                                                                        & \begin{tabular}[c]{@{}c@{}}6 Months\\ May 2019 - November 2019\end{tabular}   & 264,960                                                         \\
California                                               & 1                                                                                                         & \begin{tabular}[c]{@{}c@{}}1 Year\\  June 2014 – July 2015\end{tabular}       & 515,520       \\\hline                                                 
\end{tabular}
\caption{Number of houses and the characteristics of the PV power time series at 1 minute resolution for each location.}
\label{tab:solarData}
\end{table}

Meteorological variables are used as exogenous inputs to forecasting models to achieve more accurate PV power forecasts~\citep{Alskaif2020AEstimation, Li2014AnSystem, Liu2019ASystems}. To include the effect of weather variables on the forecasts, we collect weather data from Dark Sky~\citep{darkSky}. The weather parameters contain wind speed, temperature, dew point, cloud cover, UV index, humidity and air pressure at a 1 hour resolution. These parameters are chosen based on the importance shown in the literature with regard to PV power forecasting~\citep{Alskaif2020AEstimation}. 

\subsection{Resolutions and Horizons}

Based on our analysis of the temporal resolution and horizon in the literature (discussed in Section~\ref{sec:resolutions}), the resolution and horizon pairs listed in Table~\ref{tab:resolutions} are chosen for evaluation in this study. We consider these to be representative of the most common applications for distributed solar PV power forecasting.

The PV power data collected from Dataport is at a 1-minute resolution, and therefore, to obtain the other resolutions of interest: 5 minutes, 1 hour and 1 day, the 1 minute time series is temporally aggregated with the mean power generation of the respective time interval. This process is performed independently for all the time series obtained from the 25 households listed in Table~\ref{tab:solarData}. 
An example of the processed PV power data of one house in Austin at 1 minute, 5 minutes and 1 hour is shown in Figure~\ref{fig:resolutions}. It can be seen that in high resolutions, sudden fluctuations of the power generation can be visible in shorter time horizons, while in lower resolutions these patterns are much smoother.

To complement the resolutions of PV power data, the weather data at 1 hour resolution is pre-processed and temporally aggregated to a 1-day resolution. Next, to derive higher resolutions of 1 minute and 5 minutes, the weather variables at 1 hour resolutions are temporally disaggregated by interpolating the data for 1 minute and 5 minutes time intervals.

\begin{table}[th]
\centering
\begin{tabular}{|c|c|c|}
\hline
\textbf{Resolution} & \textbf{Horizon} & \textbf{\begin{tabular}[c]{@{}c@{}}Number of steps\\  $(horizon \div resolution)$\end{tabular}} \\ \hline
1 day               & 3 days           & 3                                                                    \\
1 hour              & 1 day            & 24                                                                   \\
5 minutes           & 1 hour           & 12                                                                   \\
1 minute            & 5 minutes        & 5   \\\hline                                                                
\end{tabular}
\caption{Resolution of the data and the respective horizon of forecasts}
\label{tab:resolutions}
\end{table}

\subsection{Choice of Evaluation Metric}

Many accuracy metrics have been studied in the literature to measure the performance of a forecasting method. Mean Absolute Error (MAE), Root Mean Squared Error (RMSE), Normalized Root Mean Squared Error (NRMSE), Mean Bias Error (MBE), Normalized Mean Bias Error (NMBE), and Mean Absolute Percentage Error (MAPE) have been commonly used in the solar PV power forecasting literature~\citep{Antonanzas2016ReviewForecasting, Zhang2015AForecasting}. However, RMSE, MBE and MAE are scale-dependent metrics, meaning that they are relative to the scale/unit of the data. When comparing the accuracy of models across data sets that have different scales/units, these measures are not very useful. On the other hand, NRMSE, NMBE, and MAPE are metrics that are scale-free and therefore helpful when comparing models across multiple data sets. However, MAPE has the disadvantage of (i) being infinite/undefined in scenarios where the actual values are zero and (ii) having a skewed error distribution when the actual values are closer to zero. An extensive review on the accuracy measures for forecasting was conducted by \cite{Hyndman2006AnotherAccuracy}. 

% In contrast to the above metrics, 
Mean Absolute Scaled Error (MASE) is a metric that compares a model's performance with respect to the performance of a naive/persistence forecast model (a model that assumes the forecasts are the same as the past observed values)~\citep{Hyndman2006AnotherAccuracy}. It is also scale-independent and therefore suitable to compare data sets that have different scales/units. MASE is calculated as shown below, where $h$ denotes the forecast horizon, $n$ is the number of in-sample (training) data points, $y_t$ and $f_t$ are the actual value and the forecast at time $t$ and $m$ is the seasonal frequency (e.g., for solar PV power generation data at an hourly resolution, the seasonal frequency will be 24).

\begin{equation}
     MASE = \frac{1}{h}\frac{\sum_{t=n+1}^{t=n+h}|y_t - f_t|}{\frac{1}{(n-m)}\sum_{t=m+1}^{t=n} |y_t -  y_{t-m}|}
\end{equation}

MASE also provides error interpretability: a forecasting model having a MASE value less than 1 performs better than a naive method on average, while a value greater than 1 indicates that the method performs poorly compared to a naive forecast computed on the training data. Considering these factors, we use MASE as our primary performance metric in the remainder of this paper. 

\section{Forecast Combination Approach}\label{sec:method}

Our forecast combination approach consists of two main parts: (i) Training base forecasters and (ii) Finding the weights that are assigned to each of these base forecasters using a given combination strategy.

\subsection{Choice of Base Forecasters}
In this study, we use a set of five base forecasters that are commonly used in many studies and practical applications. They were chosen due to their diverse underlying approaches, as diversified base forecasters complement each other during the combination process \citep{Atiya2020WhyWell}.

\vspace{0.2cm}
\noindent
\textbf{Seasonal Naive (SN):} SN (also called periodic persistence) models generate forecasts that are equivalent to the last known observations of the same time horizon. 
While this is simple forecasting approach, it is often used as a baseline to compare more advance forecasting approaches. In addition, studies have shown that SN outperforms complex forecasting models in intra-hour horizons (e.g., forecasting a few minutes ahead)~\citep{Antonanzas2016ReviewForecasting, Trapero2015Short-termRegression}. Therefore, SN is included as a base forecaster as it is particularly useful when forecasting short horizons at high resolutions.

\vspace{0.2cm}
\noindent
\textbf{Seasonal / non-seasonal Autoregressive Integrated Moving Average (SARIMA / ARIMA):} ARIMA is a statistical forecasting model in which the forecast variable is expressed using linear combinations of both the past values of that variable and past forecast errors. ARIMA models are represented with three variables $p, q, d$ where $p$ is the autoregressive component,i.e., past values, $q$ is the moving average component,i.e., past forecast errors and $d$ is the order of differencing to convert a time series to a non-stationary series. Equation \ref{eq:arima} presents an ARIMA (p,q,d) model where $y'_{t}$ is the series after differencing, $\phi, \theta$ are coefficients, $\varepsilon$ are forecast errors, and $c$ is a constant.
\begin{equation}
  y'_{t} = c + \phi_{1}y'_{t-1} + \cdots + \phi_{p}y'_{t-p}
     + \theta_{1}\varepsilon_{t-1} + \cdots + \theta_{q}\varepsilon_{t-q} + \varepsilon_{t}
\label{eq:arima}
\end{equation}

ARIMA models are extended to seasonal ARIMA (SARIMA) when seasonal patterns are apparent in the data. Using the same notation, a SARIMA model can be represented as $ARIMA$ $(p, q, d)$ $(P, Q, D)_m$ where $(p,q,d)$ is the non-seasonal component, $(P,Q,D)$ is the seasonal component and $m$ is the seasonality. When considering PV power data at different resolutions, daily seasonality (e.g., high power generation during the mid-day and low generation during morning and evening) is clearly relevant at 1-minute, 5-minute and 1-hour resolutions. However, at a 1-day resolution such seasonality is no longer relevant. Thus, we implement SARIMA models for 1-minute, 5-minute and 1-hour resolutions and an ARIMA model for a 1-day resolution.

\vspace{0.2cm}
\noindent
\textbf{Seasonal / non-seasonal Autoregressive Integrated Moving Average with Exogenous Inputs (SARIMAX / ARIMAX):} The ARIMAX model is an extension of the ARIMA model which takes into account the effect of external variables. ARIMAX models can also be extended to take seasonality into account. Similar to the ARIMA models, an ARIMAX model is built for a 1-day resolution while SARIMAX models are built in other resolutions. Weather parameters are added as external variables to these models.

\vspace{0.2cm}
\noindent
\textbf{Multiple Linear Regression (MLR):} A multiple linear regression model builds a linear relationship between the forecast variable and multiple external variables. To forecast the PV power generation, we consider weather parameters as the external variables to the MLR model. Equation~\ref{eq:1} shows the matrix representation of the variables when forecasting a horizon of length $h$. $\{x^1_{t+h}, x^2_{t+h}, \dots,x^m_{t+h}\}$ are the $m$ weather parameters at time $t+h$ which are the input variables and $f_{t+h}$ indicates the forecast.

\begin{equation}
\left[
\begin{array}{cccc|c}
x^1_{t+1} & x^2_{t+1} & \dots & x^m_{t+1} & f_{t+1} \\
\vdots & \vdots & \ddots & \vdots & \vdots \\
x^1_{t+h} & x^2_{t+h} & \dots & x^m_{t+h} & f_{t+h}
\end{array}
\right]
\label{eq:1}
\end{equation}

\vspace{0.2cm}
\noindent
\textbf{Support Vector Regression (SVR):} A Support Vector Machine (SVM) is a type of supervised machine learning algorithm commonly used for classification problems~\citep{Liu2019ASystems}. The objective of this algorithm is to find the optimal decision boundary that separates the data classes. When these models are applied for regression problems like time series forecasting, it is known as SVR. In a forecasting scenario, the SVR model tries to find the optimal line (with a given set of external features) that helps in predicting the value of interest. Similar to the MLR model, weather parameters are added as external variables to the SVR model.

\subsection{PSO-Based Forecast Combination Approach}

The objective of the forecast combination approach is to find a set of optimal weights to combine the forecasts produced by the base forecasters. For this purpose, we hold out $k$ samples from the training data. Given a time horizon of $h$, the forecasts produced by $n$ base forecasters for these samples can be represented by $F = [F_1, F_2, \dots, F_k]$, where $F_k$ is an $h \times n$ forecast matrix for the $k^{th}$ sample as shown in Equation~\ref{eq:2}, and each $f_{ij}$ represents the forecast produced by the $j^{th}$ base forecaster at time point $i$.

\begin{equation}
    F_k = \begin{bmatrix}
  f_{11} & \dots & f_{1n}\\ 
  \vdots & \ddots & \vdots \\
  f_{h1} & \dots & f_{hn}
\end{bmatrix}_{h \times n}
\label{eq:2}
\end{equation} 
\newline

The forecasts $F = [F_1, F_2, \dots, F_k]$ produced by $n$ base forecasters for the $k$ hold out samples are used to find the weights from the forecast combination methods. A MASE value can be calculated for each sample $1,2,\dots,k$. The objective of a combination strategy is to minimise the mean MASE of the $k$ samples and find a set of values $W = \{w_1, w_2,\dots, w_i, \dots, w_n\}$ where $w_i$ represents the weight given to the $i^{th}$ base forecaster. The forecast produced by the combination strategy can be represented as $F' = [F'_1, F'_2, \dots, F'_k]$, where $F'_k$ is a $h \times 1$ forecast matrix for the $k^{th}$ sample as shown in Equation \ref{eq:comb}.

\begin{equation}
    F'_k = \begin{bmatrix}
  f_{11} & \dots & f_{1n}\\ 
  \vdots & \ddots & \vdots \\
  f_{h1} & \dots & f_{hn}
\end{bmatrix}_{h \times n} \cdot \begin{bmatrix}
  w_{1}\\ 
  \vdots\\
  w_{n}
\end{bmatrix}_{n \times 1}
\label{eq:comb}
\end{equation} 
\newline

To find the set of weights $W$, we introduce a combination approach based on Particle Swarm Optimization (PSO). PSO is a population-based search optimization algorithm~\citep{Kennedy1995ParticleOptimization}. It starts with the random initialization of a population of particles, i.e., swarms in the search space, and then finds the global optimum by adjusting the path of particles towards two directions in each iteration: the individual best location of the individual particle, and the location of the best particle in the whole population. Compared to other optimization algorithms PSO has the ability to quickly converge to a reasonable solution at a lower computational cost and is therefore used in many applications including forecasting studies \citep{Dong2015AIrradiance, Ghimire2019Wavelet-basedPrediction, Kuranga2022AForecasting, Huang2022MultivariateForecasting}.

In PSO, a particle $i$ in a $d$ dimensional search space is represented with two $d$ dimensional vectors: position vector ($X_i$) and velocity vector ($V_i$). In each iteration of the algorithm, a particle's path is adjusted based on the position and velocity vectors as shown below:

\begin{equation}
    v_{i,d}^{t+1} = \hat{w} \times v_{i,d}^{t} + c_1 r_1 \times (p_{i,d}^{best} - x_{i, d}^{t}) + c_2 r_2 \times (g_{d}^{best} - x_{i, d}^{t})
\end{equation}

\begin{equation}
     x_{i, d}^{t+1} =  x_{i, d}^{t} + v_{i,d}^{t+1}
\end{equation}

where $v_{i,d}^{t}$ and $x_{i, d}^{t}$ are the current velocity and position of the $d^{th}$ dimension in the $t^{th}$ iteration. $\hat{w}$ is the inertia weight, $c_1$ and $c_2$ are acceleration constants, $r_1$ and $r_2$ are uniform random numbers that lie in the range of $[0,1]$, $p^{best}$ is the local best position of the particle and $g^{best}$ is the global best position found so far.

In our application, each particle in PSO represents a potential solution for the weight vector $W$ that is used to combine the forecasts produced by base forecasters. Since we use five base forecasters the search space is of 5 dimensions, and a particle's position and velocity will be represented by five coordinates, corresponding to the weights given to each of the base forecasters. The following steps describe the PSO-based forecast combination approach:
\begin{enumerate}
    \item Produce forecasts from the base forecasters for the $k$ hold-out samples. Each sample corresponds to a $h \times n$ matrix as shown in Equation \ref{eq:2}.
    \item Initialize the parameters $\hat{w}, c_1, c_2$ (these parameters are tuned automatically and is explained in Section \ref{sec:hyper-parameters}) and number of particles in the search space. Randomly generate the initial position and velocity vectors for each particle. 
    \item Define the objective function. We define the objective function to minimise the mean MASE across all $k$ samples. This allows to find a set of weights that are generalisable across multiple forecast samples.
    \item  Generate the combined forecasts based on the particle's position $X_i$. The weight matrix $W$ corresponds to the position vector of a particle $X_i$. Therefore, the combined forecasts can be generated as shown in Equation \ref{eq:comb}. Calculate the fitness, i.e., mean MASE across the samples for each particle.
    \item Based on the mean MASE values for each particle, update the $p^{best}$ and $g^{best}$ as required.
    \item If the termination condition, i.e., the maximum number of iterations is satisfied stop the iteration and return the best position found. Otherwise move to Step 4. 
\end{enumerate}

Several conditions can be imposed when finding these weights. For example:
\begin{enumerate}
    \item $w_i \in [0,1]$
    \item $w_i \in [0,1]$ and $\sum_{i=1}^{i=n}w_i = 1$ 
\end{enumerate}
As discussed in Section~\ref{sec:forecastingtechniques}, the most commonly used condition in forecasting studies is to consider the second case, where the weights are within $[0, 1]$ and sum to one (a convex combination)~\citep{Atiya2020WhyWell}. However, it is possible for the convexity conditions to lead to a sub-optimal solution of the weights during the optimisation process. Therefore, we explore how the above conditions imposed on the weights may affect the forecast accuracy for PV power data.
We study three combination strategies with PSO which are inherently different from one another based on the conditions imposed on the weights. 

\vspace{0.2cm}
\noindent
\textbf{Combination Strategy 1 - PSO-unconstrained: }In the first combination strategy we do not impose any conditions in finding the weights for the base forecasters, i.e., the search space of the PSO algorithm is not limited to a certain boundary. 

\vspace{0.2cm}
\noindent
\textbf{Combination Strategy 2 - PSO [0,1]: }The second combination strategy sets a boundary of $[0,1]$ on the search space of the PSO. This will allow the optimization method to find the optimum weight combination such that the individual weights of the base forecasters are always non-negative and less than one. 

\vspace{0.2cm}
\noindent
\textbf{Combination Strategy 3 - PSO-convex: }The third combination strategy will derive the weights found through the second combination strategy and normalize the weight combination to sum up to a value of one.

\vspace{0.2cm}

\subsubsection{Other Forecast Combination Methods}
Two additional existing combination approaches are implemented for comparison with the PSO-based forecast combination approach.

\vspace{0.2cm}
\noindent
\textbf{Averaging: }Averaging the forecasts of the base forecasters is the most simple and commonly employed combination approach~\citep{Genre2013CombiningAverage}. In this approach, the forecasts of all base forecasters are averaged to derive the combined forecast.

\vspace{0.2cm}
\noindent
\textbf{Recursive Ensemble (RE): } The recursive ensemble is a recently introduced forecast combination for PV power forecasting \citep{Liu2019ASystems}. In this approach, the optimal weights are found in an iterative approach where in each iteration the forecasts of the worst model (selected based on the error) is always replaced by the average forecasts of all other base forecasters. This iterative approach is continued until a defined threshold value is met. Finally, the weight combination with the minimum error is selected. To find the weights for the base forecasters using this approach, we tune the threshold value (as explained in Section \ref{sec:hyper-parameters}) using the forecasts produced by the base forecasters for the $k$ hold out samples. 

\subsection{Hyper-parameter Selection}
\label{sec:hyper-parameters}
The above explained forecasting approaches have various hyper-parameters that need to be tuned. The two time series models have three parameters $p, q, d$ for non-seasonal models while there are six parameters $p,q, d$ and $P, Q, D$ for seasonal models. These parameters are tuned using the \textit{Auto Arima}~\citep{hyndman2018forecasting} method that finds the best SARIMA/ARIMA model through: (i) unit root tests to determine $d, D$, (ii) minimisation of the corrected Akaike information criterion and maximum likelihood expectation~\citep{hyndman2018forecasting}. The hyper-parameters of SVR, PSO (all PSO-based combination strategies) and RE are tuned using a \textit{Random Search}~\citep{bergstra2012random} method which finds the best hyper-parameter combination by going through multiple iterations and a random hyper-parameter combination is selected for evaluation in each iteration. \textit{Random Search} is chosen here as it is fast and efficient in comparison to other hyper-parameter selection methods (e.g. manual and grid search) when dealing with large data sets~\citep{bergstra2012random}.

\begin{table}[th]
\centering
\small
\begin{tabular}{|l|l|l|}
\hline
\textbf{Method}                  & \textbf{Hyper-parameters}                                                                                                                                                                                        & \textbf{Tuning method} \\ \hline
\begin{tabular}[c]{@{}l@{}}(S)ARIMA\\ (S)ARIMAX\end{tabular}                    & \begin{tabular}[c]{@{}l@{}}p - non seasonal AR term\\ q - non seasonal MA term\\ d - non seasonal difference\\ P - seasonal AR term\\ Q - seasonal MA term\\ D - seasonal difference\end{tabular}    & Auto ARIMA                    \\ \hline
SVR                     & \begin{tabular}[c]{@{}l@{}} gamma\\ epsilon \end{tabular}                                                                                                                                                                                                    & Random Search                             \\ \hline
\begin{tabular}[c]{@{}l@{}}PSO- unconstrained\\ PSO [0,1]\\ PSO- convex\end{tabular}                      & \begin{tabular}[c]{@{}l@{}}$c_1$ - cognitive force\\ $c_2$ - social force\\ $\hat{w}$ - inertia\\number of neighbours\end{tabular}                                                                   & Random Search               \\\hline
RE                      & threshold                                                                                                                                                                                               & Random Search                 \\ 
\hline
\end{tabular}
\caption{Hyper-parameters of the methods and the respective tuning method used for hyper-parameter tuning. AR: Auto Regressive, MA: Moving Average.}
\label{tab:hyperparameters}
\end{table}

\section{Experimental Analysis and Results}\label{sec:results}

\subsection{Evaluation}

\begin{figure}[!th]
    \centering
    \includegraphics[width=\textwidth]{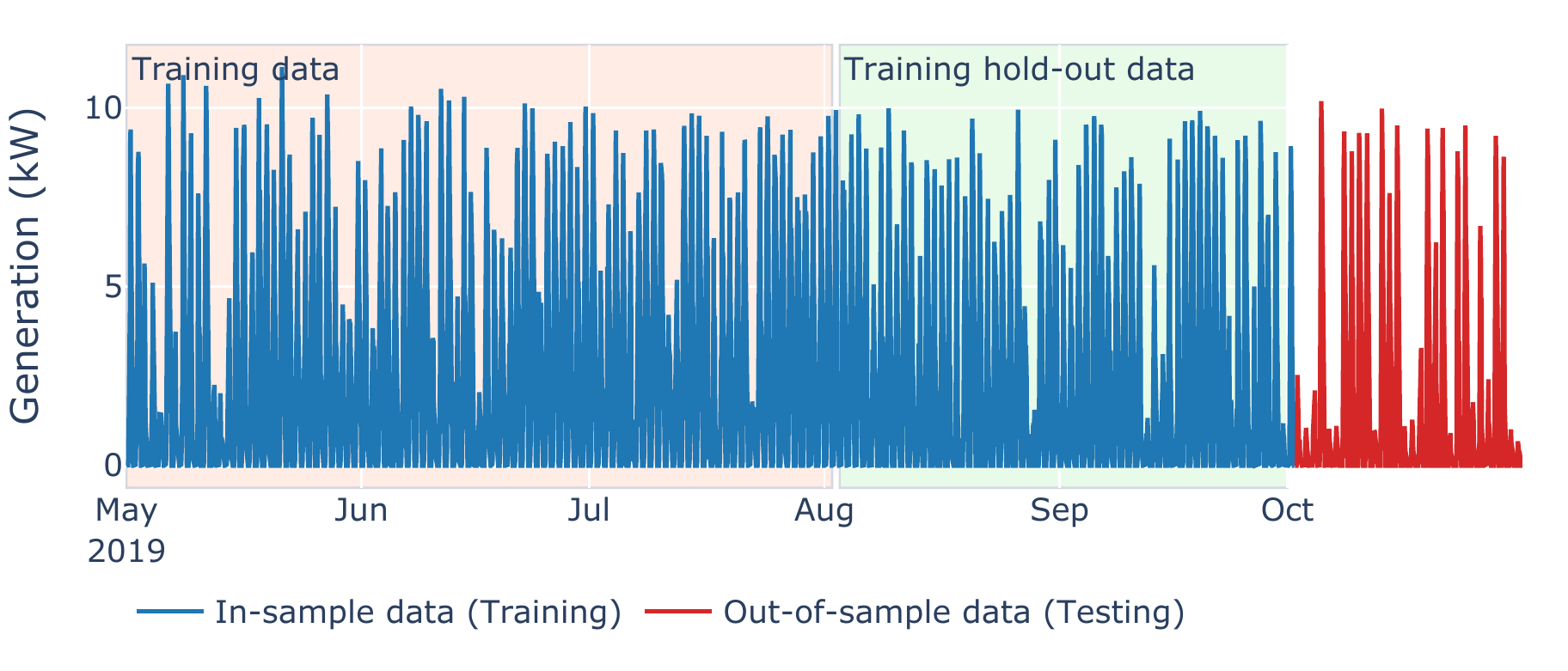}
    \caption{Illustration of splitting 1 hour resolution PV power data for one house. In-sample data is used to train the base forecasters to derive forecasts for the testing (out-of-sample) data used in the final evaluation. The Training data marked in the pink box is used to train the base forecasts to derive forecasts for the Training hold-out data (marked in the green box). The Training hold-out data is used to find the weights during the forecast combination process.}
    \label{fig:datasplit}
\end{figure}

\begin{figure}[!th]
    \centering
    \includegraphics[width=\textwidth]{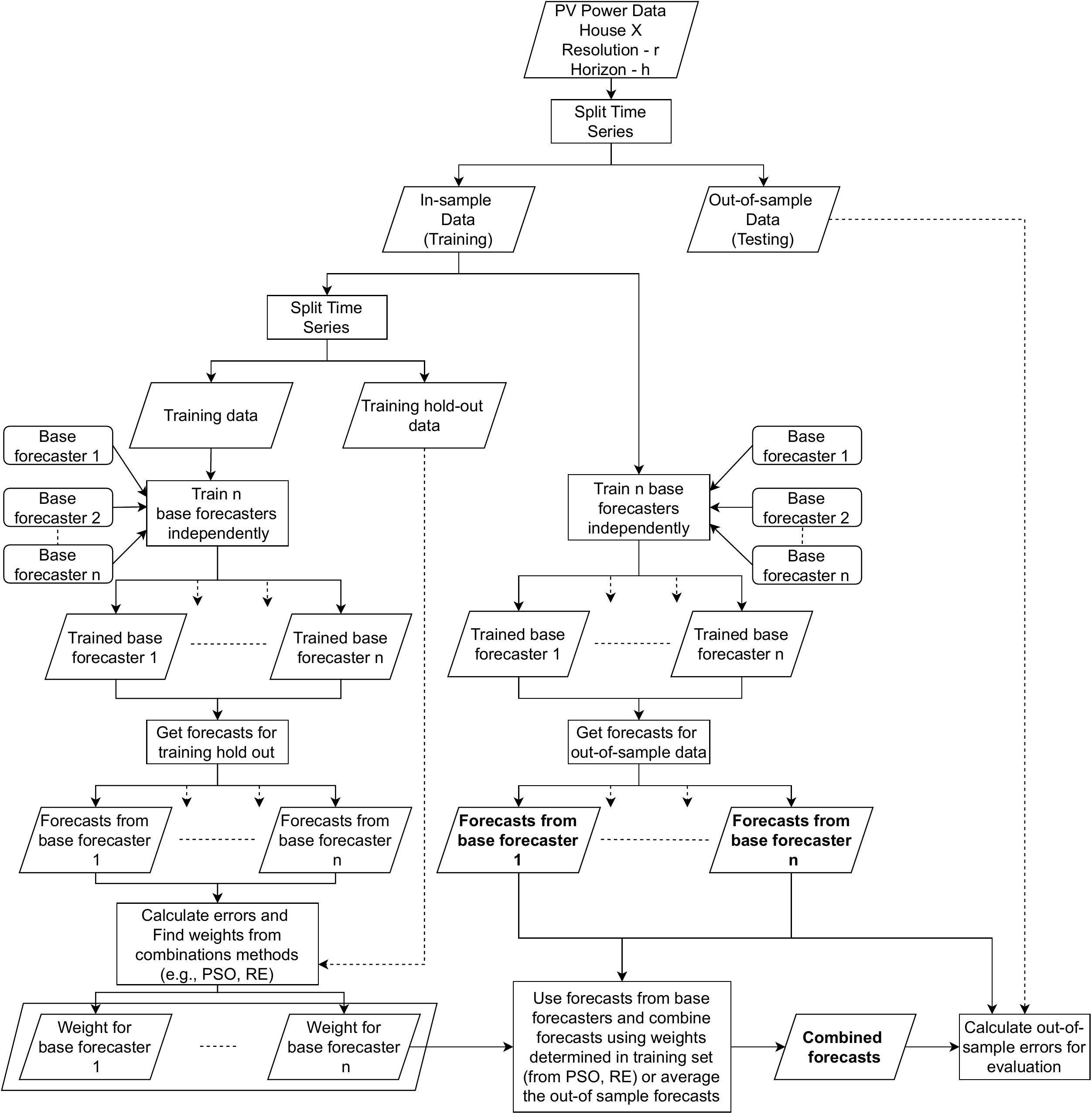}
    \caption{Illustration of the forecast combination process. The testing set samples (out-of-sample) forecasts used for evaluation are marked in boldface.}
    \label{fig:comb_process}
\end{figure}

We conduct an out-of-sample evaluation to analyse the performance of base forecasters and forecast combinations. Figure~\ref{fig:datasplit} shows an example of how we partition the data for a single house. The last month of every time series is withheld for evaluation as the testing set (out-of-sample data). The test samples include days having different weather conditions (e.g., cloudy, sunny days) as they are derived from one month worth of data. For the out-of-sample evaluation, the base forecasters are trained using the rest of the data (in-sample data) and the forecasts are derived for the test samples. As explained in Section~\ref{sec:method}, to find the weights using the combination strategies, we hold out samples, i.e., training hold out data, from in-sample data -- in this case two months. Figure~\ref{fig:comb_process} shows a graphical representation of the forecast combination process. The weights determined from the combination strategies are applied to the forecasts produced by the base forecasters for the test samples to obtain the combined forecasts for evaluation. The evaluation process is conducted independently for all 25 households. 

The MASE for each sample in the testing set is calculated independently. Finally, the mean MASE values across all test samples are calculated for all the houses separately to determine the performance of the forecasting approaches per household. To consolidate the errors across all houses, the median of this mean MASE distribution across households is used as the final error. The median MASE is reported as it is robust to outliers when dealing with a relatively small error distribution of 25 time series. A lower median in the distribution reflects that a particular approach has reduced the error for most houses and therefore has a better forecast performance. 

As previously shown in Table~\ref{tab:solarData}, the data set consists of time series for three locations across three different time periods. Therefore, our evaluation process compares the performance of the forecasting approaches: (a) when trained with different amounts of data and (b) in different geographical locations. For each time series, the model training and evaluation time periods are kept consistent across all horizons and resolutions and across all models for a fair evaluation. The only exception was made for ARIMA models at high resolutions due to their extensive training times, as justified in \ref{app:arima}. For SARIMA models we use 25 days of recent data to forecast a 5 minute horizon with minutely data, and 14 days of recent data to forecast a 1 hour horizon at a 5 minute resolution. All experiments were conducted using the High Performance Computing System at the University of Melbourne~\citep{lafayette2016spartan}.

\subsection{Results and Discussion}
\label{sec:resultsdiscussion}

\begin{figure*}[]
     \centering
     \begin{subfigure}[t]{\textwidth}
         \centering
         \includegraphics[width=\textwidth]{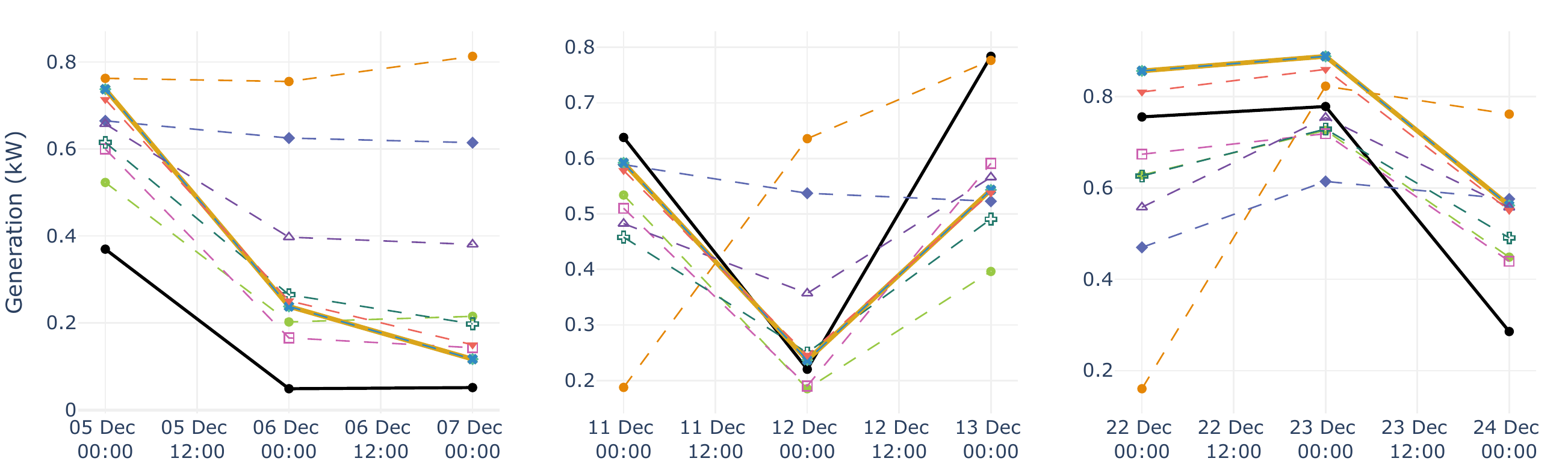}
         \caption{3 day horizon at 1 day resolution}
         \label{fig:3D_forecasts}
     \end{subfigure}
     ~
     \begin{subfigure}[t]{\textwidth}
         \centering
         \includegraphics[width=\textwidth]{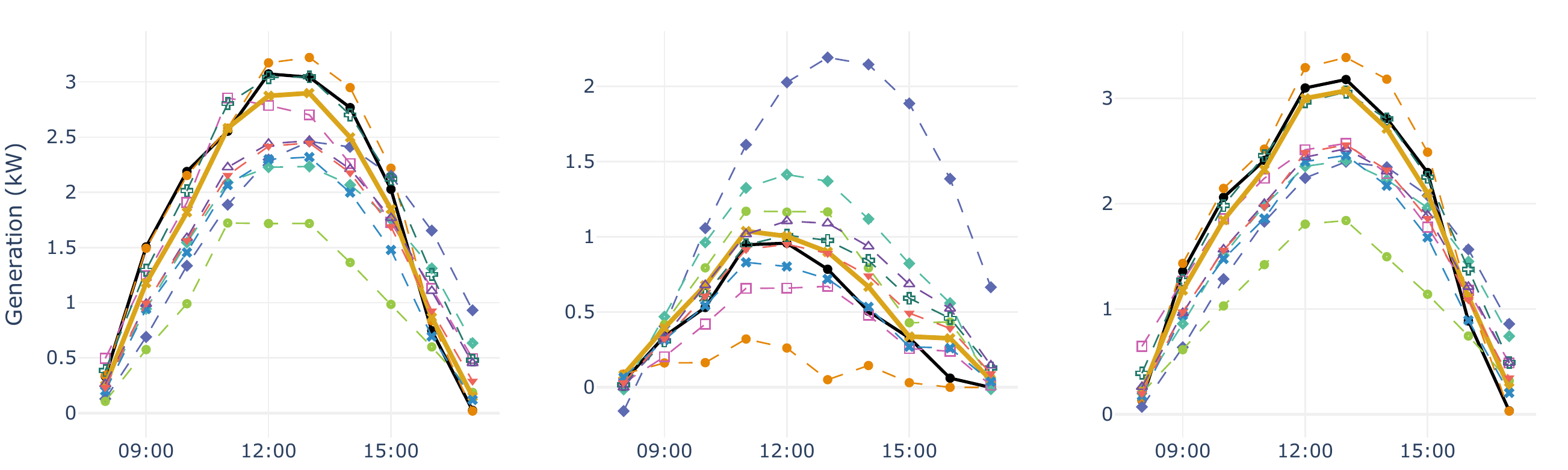}
         \caption{1 day horizon at 1 hour resolution}
         \label{fig:1D_forecasts}
     \end{subfigure}
     ~
     \begin{subfigure}[t]{\textwidth}
         \centering
         \includegraphics[width=\textwidth]{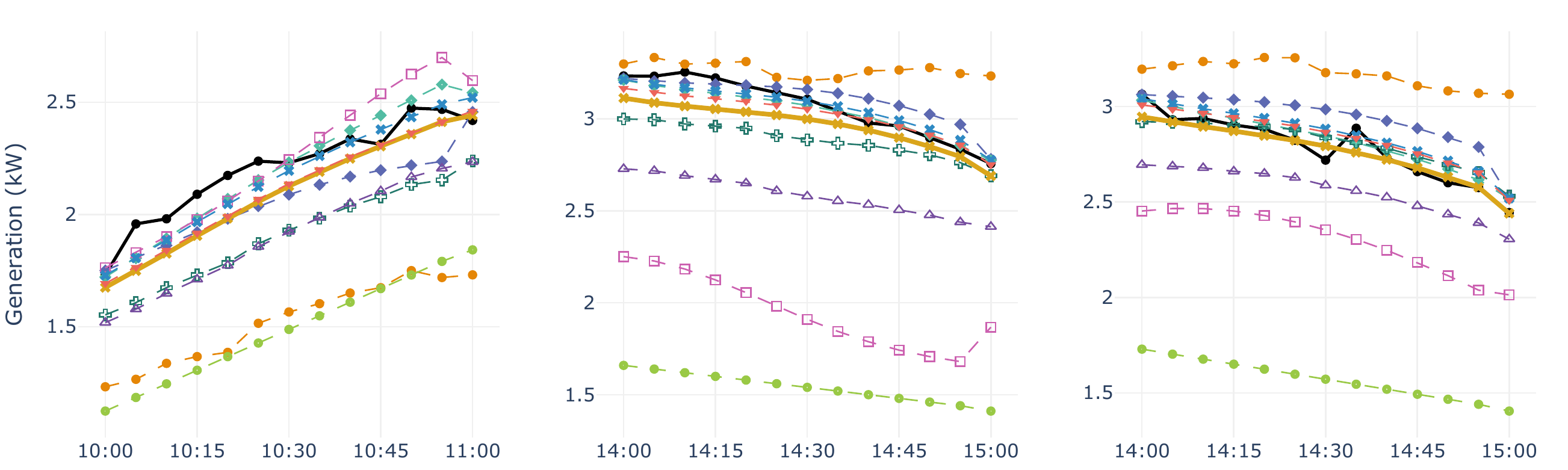}
         \caption{1 hour horizon at 5 minutes resolution}
         \label{fig:1H_forecasts}
     \end{subfigure}
     ~
     \begin{subfigure}[t]{\textwidth}
         \centering
         \includegraphics[width=\textwidth]{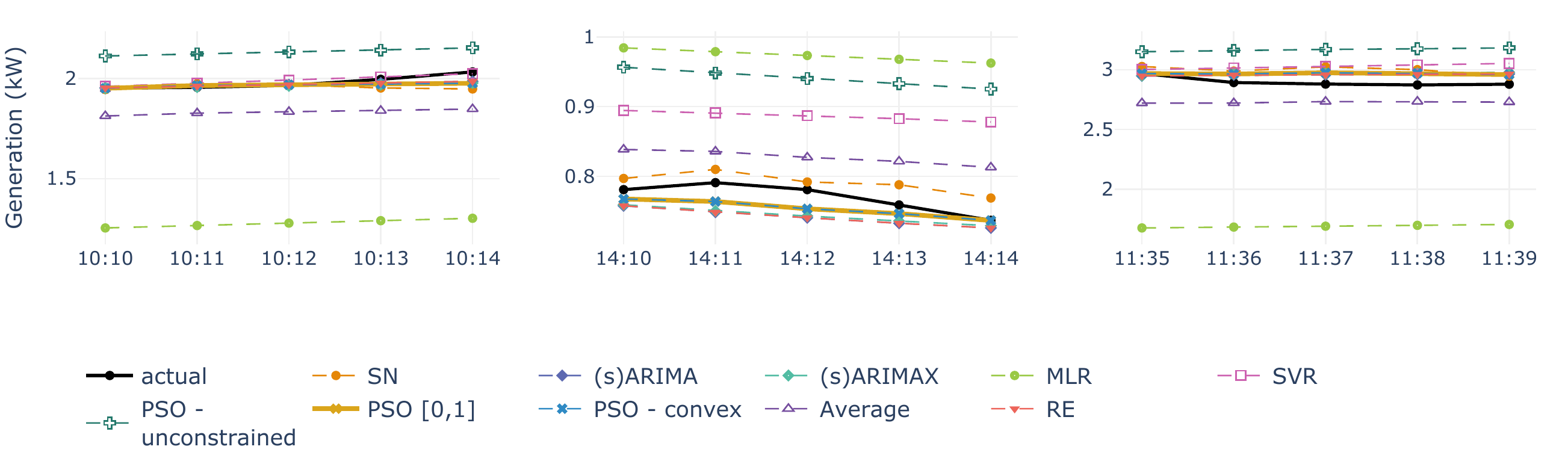}
         \caption{5 minutes horizon at 1 minute resolution}
         \label{fig:5min_forecasts}
     \end{subfigure}
    \caption{Forecasts produced by the forecasting approaches for three samples in the testing set (out-of-sample data) of one house.}
    \label{fig:forecasts}
\end{figure*}

\begin{figure*}[!h]
     \centering
     \begin{subfigure}[t]{\textwidth}
         \centering
         \includegraphics[width=\textwidth]{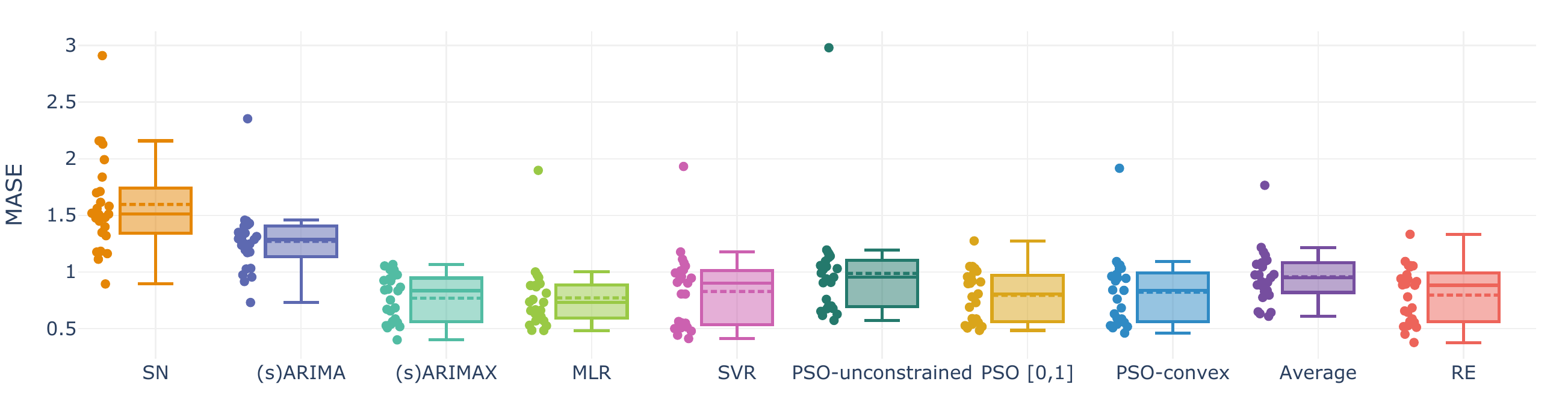}
         \caption{3 day horizon at 1 day resolution}
         \label{fig:3D_errors}
     \end{subfigure}
     ~
     \begin{subfigure}[t]{\textwidth}
         \centering
         \includegraphics[width=\textwidth]{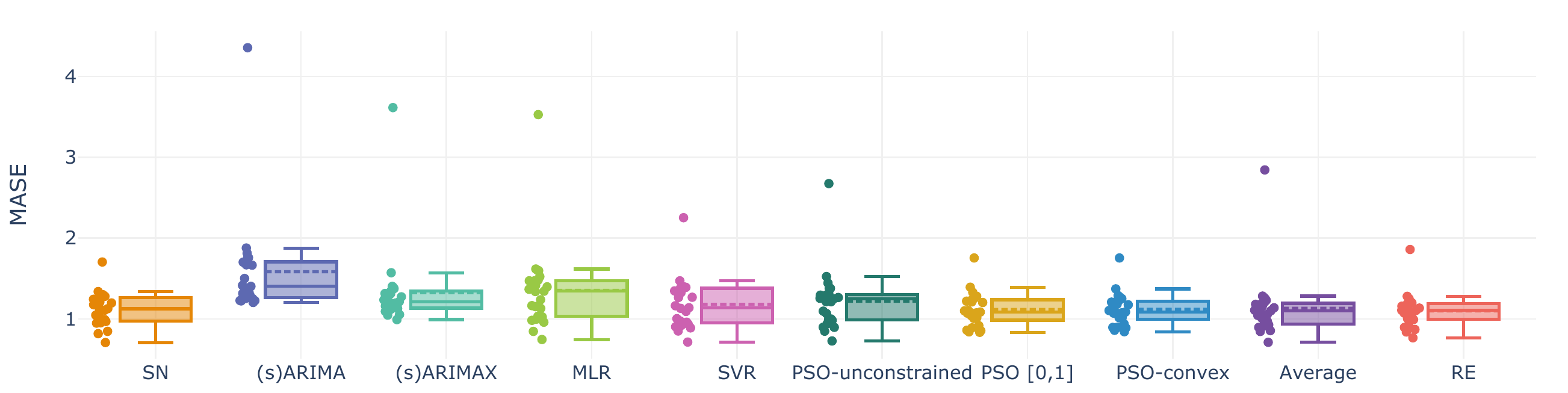}
         \caption{1 day horizon at 1 hour resolution}
         \label{fig:1D_errors}
     \end{subfigure}
     ~
     \begin{subfigure}[t]{\textwidth}
         \centering
         \includegraphics[width=\textwidth]{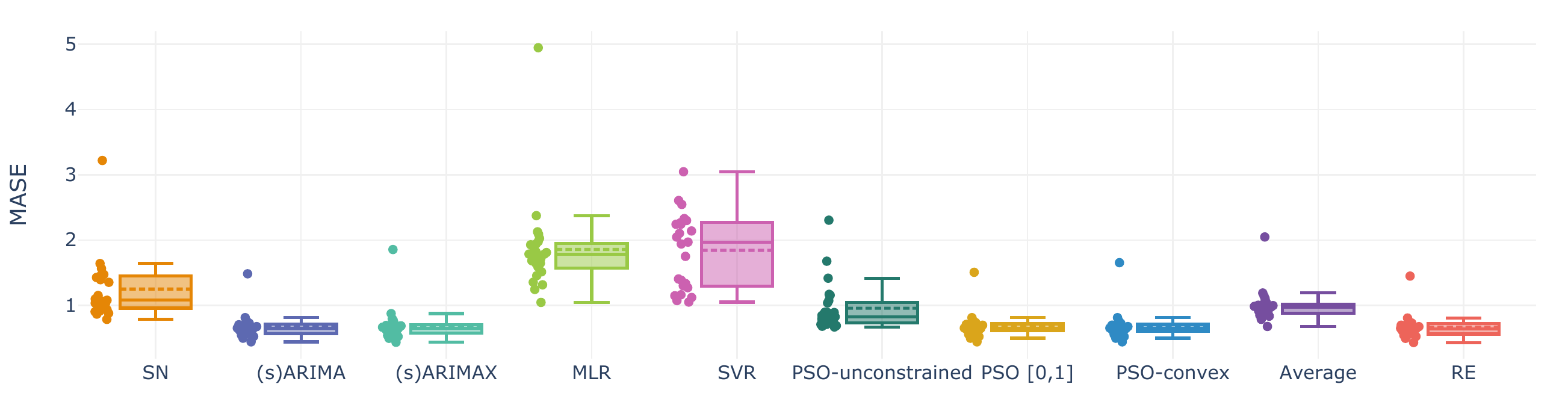}
         \caption{1 hour horizon at 5 minutes resolution}
         \label{fig:1H_errors}
     \end{subfigure}
     ~
     \begin{subfigure}[t]{\textwidth}
         \centering
         \includegraphics[width=\textwidth]{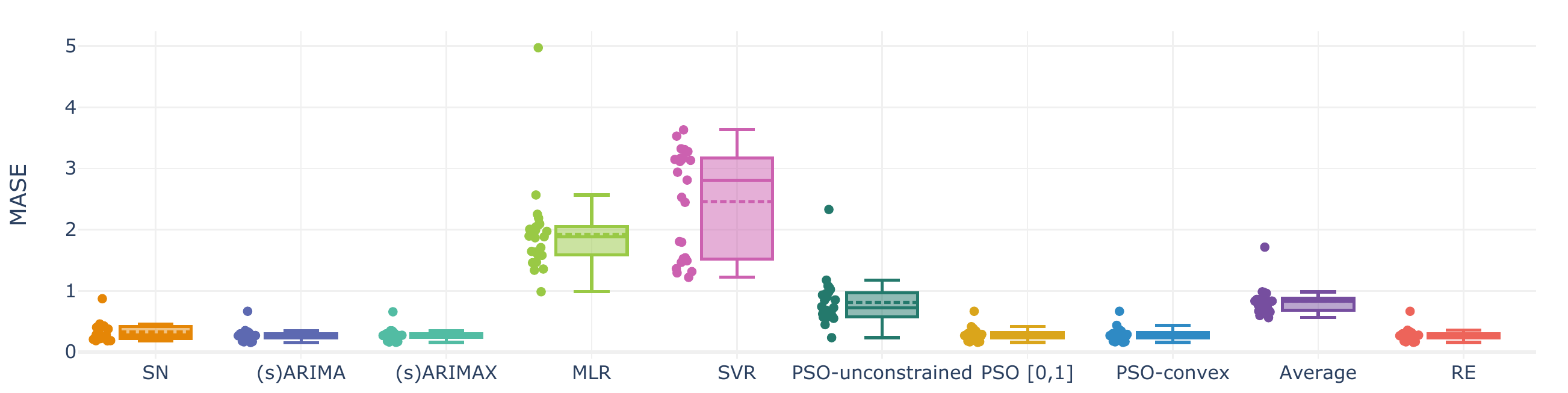}
         \caption{5 minutes horizon at 1 minute resolution}
         \label{fig:5min_errors}
     \end{subfigure}
    \caption{Mean MASE (considering all testing set samples) distribution of the 25 households.}
    \label{fig:errorDis}
\end{figure*}

An example of the forecasts for three test samples of one household at different resolution and horizon pairs are shown in Figure~\ref{fig:forecasts}. Figure~\ref{fig:errorDis} shows the mean MASE (considering all test samples) distribution of 25 households at different temporal resolution and horizon pairs considered (MASE values for each household are provided in \ref{sec:appendixB}). The median MASE of these distributions and the respective rank of the approaches are reported in Table~\ref{tab:mase_scores}. From the table, it is evident that the proposed PSO [0,1] combination strategy shows the best performance considering all resolution and horizon pairs. Furthermore, it can be seen that all other forecasting approaches have a varying performance across the different resolutions and horizons while the PSO [0,1] shows a consistent performance (always in a top 3 position). 

\begin{table}[t]
\small
\resizebox{\textwidth}{!}{\begin{tabular}{|l|r|r||r|r||r|r||r|r||r|}
\hline
\multicolumn{1}{|c|}{\multirow{3}{*}{\textbf{Method}}} & \multicolumn{8}{c|}{\textbf{Horizon (H) - Resolution (R)}}                                                                                                                                                                                                                                                                                                     & \multicolumn{1}{c|}{\multirow{3}{*}{\textbf{\begin{tabular}[c]{@{}c@{}}Final\\ Rank\end{tabular}}}} \\ \cline{2-9}
\multicolumn{1}{|c|}{}                                 & \multicolumn{2}{c||}{\begin{tabular}[c]{@{}c@{}}H - 3 days\\ R - 1 day\end{tabular}} & \multicolumn{2}{c||}{\begin{tabular}[c]{@{}c@{}}H - 1 day\\ R - 1 hour\end{tabular}} & \multicolumn{2}{c||}{\begin{tabular}[c]{@{}c@{}}H - 1 hour\\ R - 5 minutes\end{tabular}} & \multicolumn{2}{c||}{\begin{tabular}[c]{@{}c@{}}H - 5 minutes\\ R - 1 minute\end{tabular}} & \multicolumn{1}{c|}{}                                                                               \\ \cline{2-9}
\multicolumn{1}{|c|}{}                                 & \multicolumn{1}{c|}{MASE}                & \multicolumn{1}{c||}{Rank}                & \multicolumn{1}{c|}{MASE}                & \multicolumn{1}{c||}{Rank}               & \multicolumn{1}{c|}{MASE}                  & \multicolumn{1}{c||}{Rank}                  & \multicolumn{1}{c|}{MASE}                   & \multicolumn{1}{c||}{Rank}                   & \multicolumn{1}{c|}{}                                                                               \\ \hline
SN                                                     & 1.511                                    & 11                                       & 1.128                                    & 5                                       & 1.080                                       & 9                                          & 0.275                                       & 7                                           & 9.5                                                                                                 \\
(S) ARIMA                                              & 1.287                                    & 10                                       & 1.405                                    & 11                                      & 0.651                                      & 4                                          & 0.266                                       & 2.5                                           & 7                                                                                                   \\
(S) ARIMAX                                             & 0.835                                    & 3                                        & 1.214                                    & 8                                       & 0.652                                      & 5                                          & 0.269                                       & 6                                           & 4.5                                                                                                 \\
MLR                                                    & 0.733                                    & 1                                        & 1.344                                    & 10                                      & 1.786                                      & 10                                         & 1.884                                       & 10                                          & 8                                                                                                   \\
SVR                                                    & 0.903                                    & 6                                        & 1.140                                    & 7                                       & 1.970                                       & 11                                         & 2.811                                       & 11                                          & 11                                                                                                  \\
PSO - unconstrained                                                    & 0.954                                    & 8                                        & 1.251                                    & 9                                       & 0.826                                      & 7                                          & 0.725                                       & 8                                           & 9.5                                                                                                 \\
PSO {[}0,1{]}                                          & 0.807                                    & 2                                        & 1.086                                    & 1                                       & 0.648                                      & 3                                          & 0.267                                       & 4.5                                         & \textbf{1}                                                                                          \\
PSO - convex                                           & 0.838                                    & 4                                        & 1.087                                    & 2                                       & 0.647                                      & 2                                          & 0.267                                       & 4.5                                         & \textbf{2.5}                                                                                        \\
Average                                                & 0.951                                    & 7                                        & 1.102                                    & 3                                       & 0.974                                      & 8                                          & 0.832                                       & 9                                           & 6                                                                                                   \\
RE                                                     & 0.883                                    & 5                                        & 1.106                                    & 4                                       & 0.644                                      & 1                                          & 0.266                                       & 2.5                                         & \textbf{2.5}                                                                                        \\\hline
% LightGBM                                               & 1.132                                    & 9                                        & 1.129                                    & 6                                       & 0.700                                        & 6                                          & 0.255                                       & 1                                        & 4.5       \\\hline
\end{tabular}}
\caption{Median MASE across the 25 households, respective rank of the methods at different resolutions and horizons and the final rank of the methods across all resolutions and horizons. The top performing 3 approaches considering all horizons and resolutions are marked in boldface.}
\label{tab:mase_scores}
\end{table}

\begin{figure}[]
    \centering
    \includegraphics[width=\textwidth]{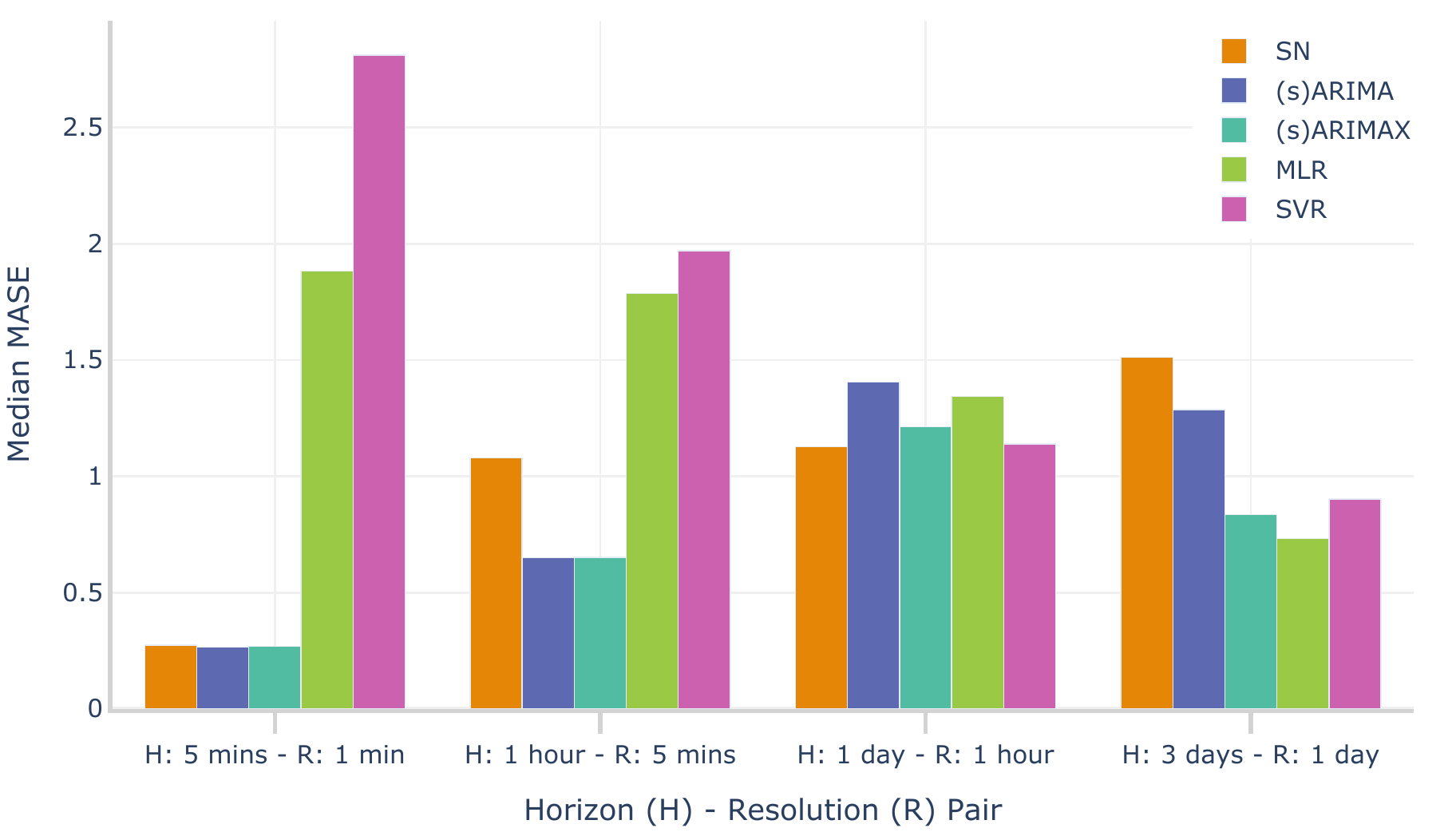}
    \caption{Median MASE values of the base forecasters (shown in Table \ref{tab:mase_scores}) across the 25 households at different temporal horizon and resolution pairs studied. The horizon and resolution pairs are displayed as H: horizon - R: resolution (e.g., H: 5 mins - R: 1 min indicate the forecast horizon is 5 minutes and the resolution is 1 minute.).}
    \label{fig:masegraph}
\end{figure}

When considering only the base forecasters, (S)ARIMAX has the best performance. According to Figure~\ref{fig:masegraph}, SN, (S)ARIMA and (S)ARIMAX perform well in high resolutions over shorter horizons. On the contrary, the two machine learning models perform poorly and also have a more inconsistent performance across the households (Figure~\ref{fig:errorDis}). However, the MLR model outperforms all other approaches at a 3 days horizon over a daily resolution while its performance significantly decreases at all other temporal dimensions. 

It can also be noted that all combination methods (except PSO-unconstrained and Average) outperform all other models. Comparing PSO [0,1] and the RE, PSO [0,1] reduces the median MASE on average by 3.81\% compared to (S)ARIMAX, while RE has only reduced the error by 1.37\%. Furthermore, it can be seen that the Averaging approach performs well at a 1 day horizon over 1 hour resolution, but has a significantly poorer performance in all other resolution and horizon pairs. This can be explained by the MASE of the base forecasters shown in Figure~\ref{fig:masegraph}. The differences in MASE between the base forecasters is relatively low at this resolution and horizon, suggesting a competitive performance among the base forecasters. Therefore, giving an equal weight to all base forecasters by averaging their forecasts when the base forecasters have a competitive performance improves the overall forecast performance. However, when the base forecasters are not competitive with each other, i.e., when the forecasts produced by base forecaster have widely varying MASE values, treating all base forecasters equally will degrade the overall forecast accuracy.  

It can be observed that all base forecasters and combinations have a MASE value greater than 1, indicating a poor forecast performance for the horizon of 1 day at a 1-hour resolution compared to all other resolution and horizon pairs. This resolution and horizon pair has the highest number of time points to forecast compared to all other resolutions and horizons (there are 24 time points in this resolution and horizon pair, while all others have less than half of this number of time points). Therefore, we anticipate that the poor forecast performance in the 1-day horizon at a 1-hour resolution may be caused by higher uncertainty associated with predicting a larger number of time points into the future.

\begin{figure*}[]
     \centering
     \begin{subfigure}[t]{0.485\textwidth}
         \centering
         \includegraphics[width=\textwidth]{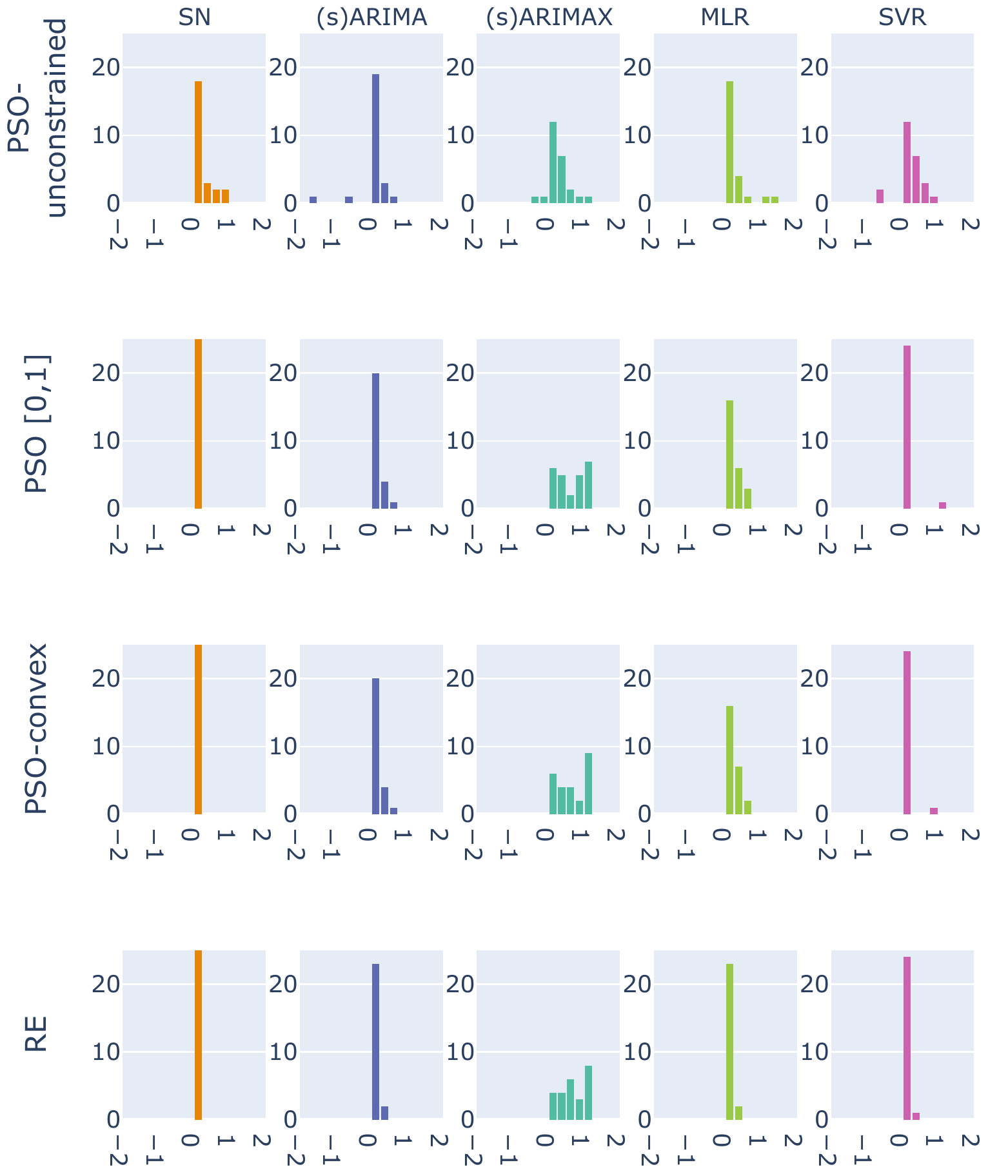}
         \caption{3 day horizon at 1 day resolution}
         \label{fig:3D_weights}
     \end{subfigure}
     ~
     \begin{subfigure}[t]{0.485\textwidth}
         \centering
         \includegraphics[width=\textwidth]{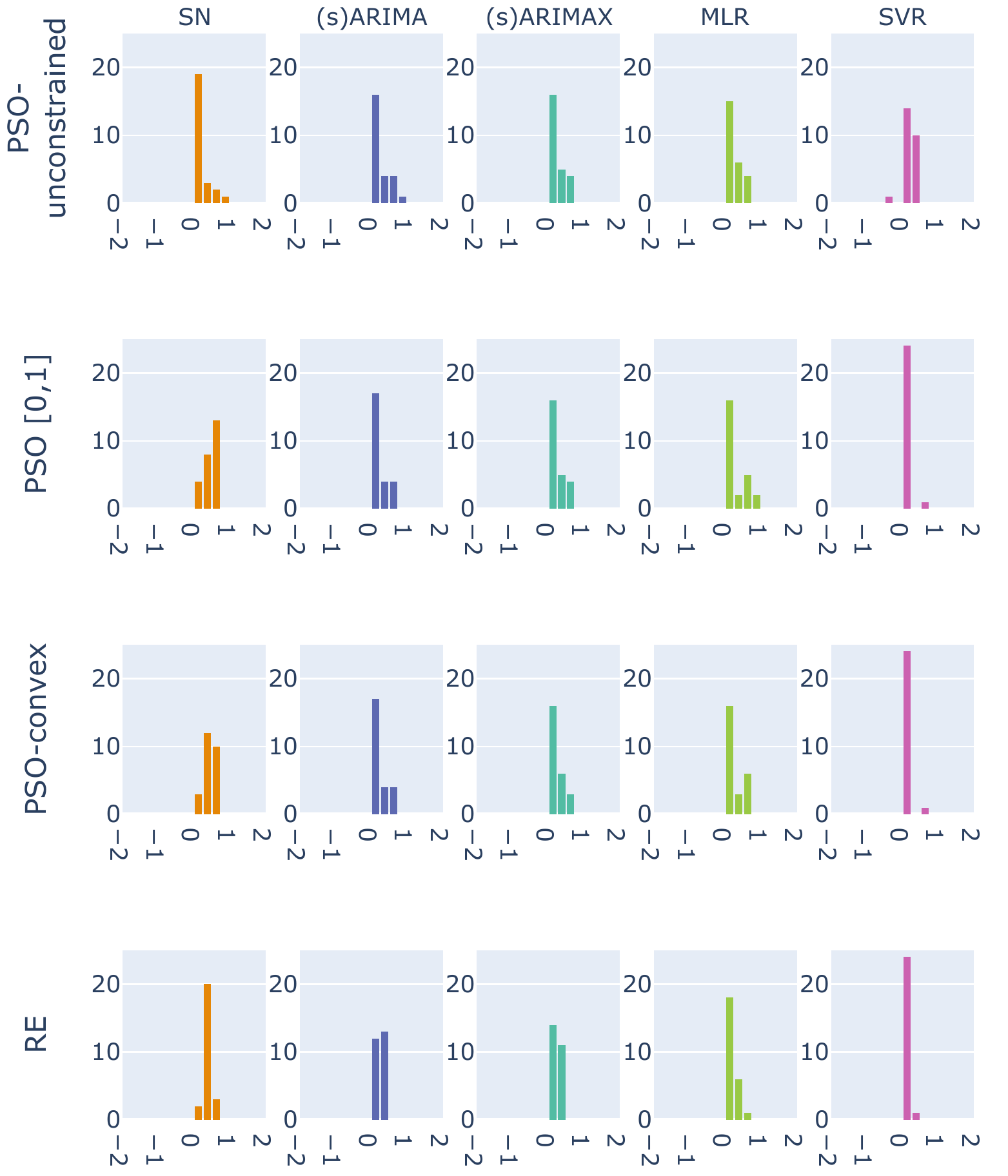}
         \caption{1 day horizon at 1 hour resolution}
         \label{fig:1D_weights}
     \end{subfigure}
     ~
     \begin{subfigure}[t]{0.485\textwidth}
         \centering
         \includegraphics[width=\textwidth]{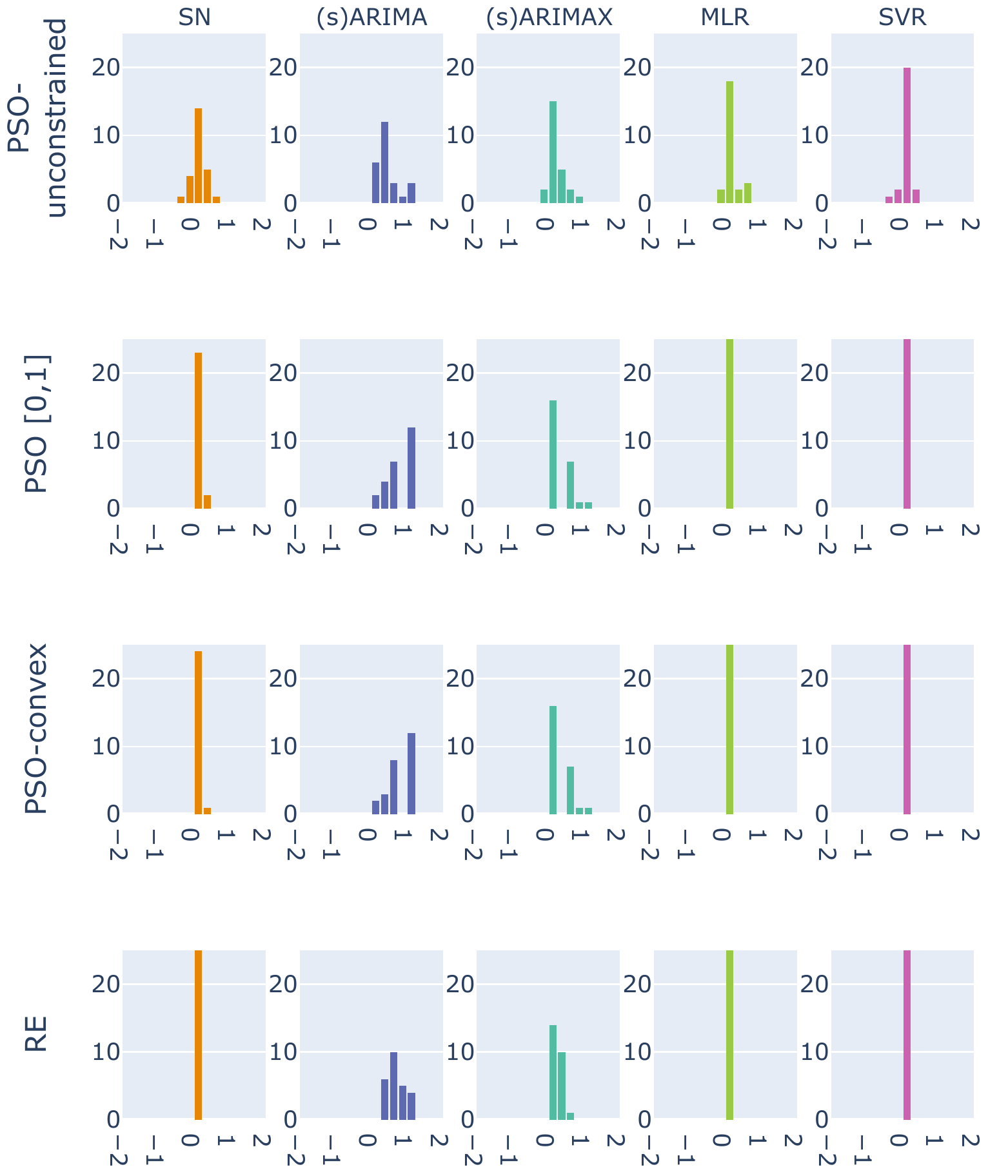}
         \caption{1 hour horizon at 5 minutes resolution}
         \label{fig:1H_weights}
     \end{subfigure}
     ~
     \begin{subfigure}[t]{0.485\textwidth}
         \centering
         \includegraphics[width=\textwidth]{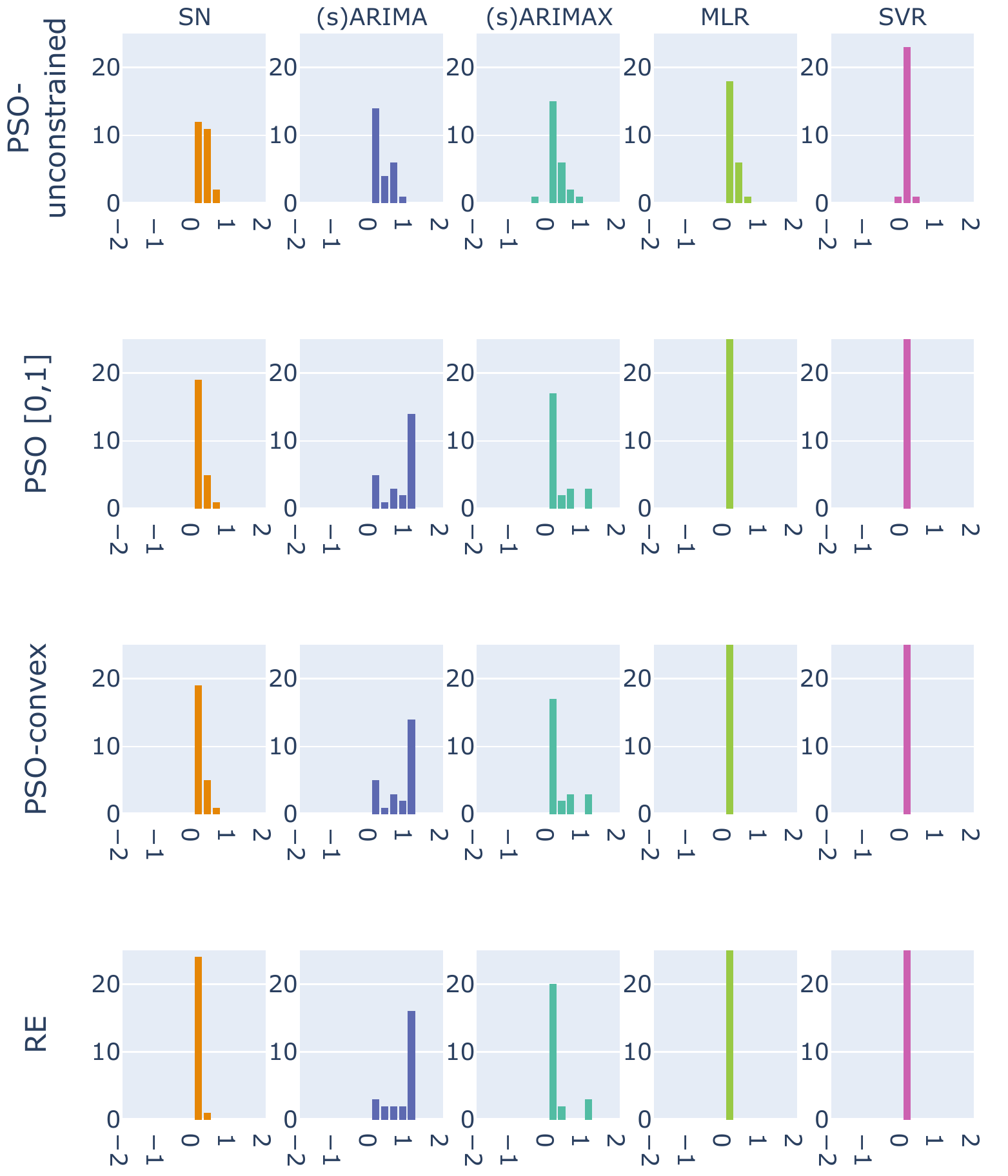}
         \caption{5 minutes horizon at 1 minute resolution}
         \label{fig:5min_weights}
     \end{subfigure}
    \caption{Histograms of the weights given by combination strategies for each base forecaster across the 25 households. Each bin in the histograms corresponds to a weight value and each bar corresponds to the number of houses a particular weight value was given to the base forecaster mentioned in the graph.}
    \label{fig:weightgraph}
\end{figure*}

Regarding the performance of the PSO-unconstrained combination strategy, it can be observed that this combination approach performs poorly across all resolution and horizon pairs. We attempt to explain this observation through the histograms shown in Figure~\ref{fig:weightgraph}, which illustrates the frequency distribution of the weights assigned to each base forecaster across the 25 households by the different combination strategies\footnote{The weight histogram for the average is not shown in the figure as it assigns the same weight value of 0.2 for all base forecasters across all households.}. PSO [0,1], PSO-convex and the RE have assigned weights to the base forecasters in similar distributions and therefore, have a competitive performance in the final results. However, the histograms of the weights obtained from the PSO-unconstrained strategy has a distribution which is much different from those approaches. The key differences in the histograms are noted at a 3-day horizon over 1-day resolution (Figure~\ref{fig:3D_weights}), PSO-unconstrained has assigned positive weights greater than 0.2 for SN, SVR while the other approaches have assigned weights between 0 and 0.2 for most houses. Also, for ARIMAX while the other approaches have given weights greater than 0.2, PSO-unconstrained has assigned both negative and positive weights and has given values between 0 and 0.2 for most houses. At a 1-day horizon over 1-hour resolution (Figure~\ref{fig:1D_weights}), all other approaches have given weights greater than 0.2 for SN across most households, while PSO-unconstrained have given weights between 0 and 0.2, at a 1-hour horizon over 5 minutes resolution (Figure~\ref{fig:1H_weights}), the key differences are observed in the weights given for ARIMA. PSO-unconstrained assigns weights between 0.2 and 0.4 for ARIMA for a majority of household while other strategies have given weights closer to 1 and at a 5-minute horizon over 1-minute resolution (Figure~\ref{fig:5min_weights}), for ARIMA the weights assigned by PSO-unconstrained for a majority of the households are between 0 and 0.2, while other approaches have given weights closer to 1. We note that having no constraints on the weights makes it difficult to find the optimal set of values for the weights to improve out-of-sample forecasts, instead may lead to an over-fitted solution for the hold out samples used in the weight finding process.

We further analyse the performance of the forecasting approaches based on the location of a household. To determine whether the location of a household has an impact on the performance of the forecasting approaches, a statistical significance test (Mann–Whitney U test) is conducted between households in Austin and New York\footnote{California is not included in the analysis because there is only one house in this location.}. Table~\ref{tab:locationwise_scores} shows the rank of the methods based on the location and the p-value of Mann–Whitney U test. The null hypothesis of the Mann–Whitney U test indicates that for two independent samples $x$, $y$ the distribution underlying sample $x$ is the same as the distribution underlying sample $y$, i.e., there is no significant difference between the two distributions and the alternative hypothesis indicates the two distributions are different. We consider a p-value less than 0.05 is required to reject the null hypothesis. Therefore, a p-value less than 0.05 indicate the method performs significantly different across the two locations - Austin and New York.

From Table~\ref{tab:locationwise_scores}, it is evident that a majority of the forecasting approaches perform significantly differently in all resolution and horizon pairs, except at a 1-day horizon over an hourly resolution. This resolution and horizon pair is also the most frequently studied horizon and resolution in PV power forecasting, as discussed in Section~\ref{sec:resolutions}. Another interesting observation is that (S)ARIMAX and SVR models have a varying performance based on the location of the household across all resolution and horizon pairs. Both these models are base forecasters and include weather information of the specific location. Therefore, the performance variability may have been caused by the different weather patterns observed in the two locations. 

\begin{sidewaystable}[]
\small
\resizebox{\textwidth}{!}{\begin{tabular}{|l|r|r|r||r|r|r||r|r|r||r|r|r|}
\hline
\multicolumn{1}{|c|}{\multirow{3}{*}{\textbf{Method}}} & \multicolumn{3}{c|}{\textbf{\begin{tabular}[c]{@{}c@{}}H - 3 days\\ R - 1 day\end{tabular}}}                                           & \multicolumn{3}{c|}{\textbf{\begin{tabular}[c]{@{}c@{}}H - 1 day\\ R - 1 hour\end{tabular}}}                                           & \multicolumn{3}{c|}{\textbf{\begin{tabular}[c]{@{}c@{}}H - 1 hour\\ R - 5 minutes\end{tabular}}}                                       & \multicolumn{3}{c|}{\textbf{\begin{tabular}[c]{@{}c@{}}H - 5 minutes\\ R - 1 minute\end{tabular}}}                                     \\ \cline{2-13} 
\multicolumn{1}{|c|}{}                                 & \multicolumn{2}{c|}{\textbf{Rank}}                                            & \multicolumn{1}{c|}{\multirow{2}{*}{\textbf{p value}}} & \multicolumn{2}{c|}{\textbf{Rank}}                                            & \multicolumn{1}{c|}{\multirow{2}{*}{\textbf{p value}}} & \multicolumn{2}{c|}{\textbf{Rank}}                                            & \multicolumn{1}{c|}{\multirow{2}{*}{\textbf{p value}}} & \multicolumn{2}{c|}{\textbf{Rank}}                                            & \multicolumn{1}{c|}{\multirow{2}{*}{\textbf{p value}}} \\ \cline{2-3} \cline{5-6} \cline{8-9} \cline{11-12}
\multicolumn{1}{|c|}{}                                 & \multicolumn{1}{l|}{\textbf{Austin}} & \multicolumn{1}{l|}{\textbf{New York}} & \multicolumn{1}{c|}{}                                  & \multicolumn{1}{l|}{\textbf{Austin}} & \multicolumn{1}{l|}{\textbf{New York}} & \multicolumn{1}{c|}{}                                  & \multicolumn{1}{l|}{\textbf{Austin}} & \multicolumn{1}{l|}{\textbf{New York}} & \multicolumn{1}{c|}{}                                  & \multicolumn{1}{l|}{\textbf{Austin}} & \multicolumn{1}{l|}{\textbf{New York}} & \multicolumn{1}{c|}{}                                  \\ \hline
SN                                                     & \multicolumn{1}{r|}{11}              & \multicolumn{1}{r|}{11}                & \textbf{\textless 0.001 *}                             & \multicolumn{1}{r|}{7}               & \multicolumn{1}{r|}{6}                 & 0.660                                                   & \multicolumn{1}{r|}{10}              & \multicolumn{1}{r|}{9}                 & \textbf{\textless 0.001 *}                             & \multicolumn{1}{r|}{7}               & \multicolumn{1}{r|}{2}                 & \textbf{\textless 0.001 *}                             \\
(S) ARIMA                                              & \multicolumn{1}{r|}{10}              & \multicolumn{1}{r|}{10}                & 0.976                                                  & \multicolumn{1}{r|}{11}              & \multicolumn{1}{r|}{8}                 & \textbf{\textless 0.001 *}                             & \multicolumn{1}{r|}{3}               & \multicolumn{1}{r|}{2}                 & \textbf{\textless 0.001 *}                             & \multicolumn{1}{r|}{2.5}             & \multicolumn{1}{r|}{5}                 & \textbf{\textless 0.001 *}                             \\
(S) ARIMAX                                             & \multicolumn{1}{r|}{4}               & \multicolumn{1}{r|}{3}                 & \textbf{\textless 0.001 *}                             & \multicolumn{1}{r|}{10}              & \multicolumn{1}{r|}{5}                 & \textbf{\textless 0.001 *}                             & \multicolumn{1}{r|}{2}               & \multicolumn{1}{r|}{3}                 & \textbf{\textless 0.001 *}                             & \multicolumn{1}{r|}{1}               & \multicolumn{1}{r|}{7}                 & \textbf{\textless 0.001 *}                             \\
MLR                                                    & \multicolumn{1}{r|}{6}               & \multicolumn{1}{r|}{1}                 & \textbf{\textless 0.001 *}                             & \multicolumn{1}{r|}{9}               & \multicolumn{1}{r|}{11}                & 0.429                                                  & \multicolumn{1}{r|}{11}              & \multicolumn{1}{r|}{10}                & 0.618                                                  & \multicolumn{1}{r|}{11}              & \multicolumn{1}{r|}{10}                & 0.661                                                  \\
SVR                                                    & \multicolumn{1}{r|}{1}               & \multicolumn{1}{r|}{6}                 & \textbf{\textless 0.001 *}                             & \multicolumn{1}{r|}{2}               & \multicolumn{1}{r|}{10}                & \textbf{\textless 0.001 *}                             & \multicolumn{1}{r|}{9}               & \multicolumn{1}{r|}{11}                & \textbf{\textless 0.001 *}                             & \multicolumn{1}{r|}{10}              & \multicolumn{1}{r|}{11}                & \textbf{\textless 0.001 *}                             \\
PSO -   unconstrained                                  & \multicolumn{1}{r|}{7}               & \multicolumn{1}{r|}{7}                 & \textbf{0.005 *}                                       & \multicolumn{1}{r|}{8}               & \multicolumn{1}{r|}{7}                 & 0.333                                                  & \multicolumn{1}{r|}{7}               & \multicolumn{1}{r|}{7}                 & 0.792                                                  & \multicolumn{1}{r|}{8}               & \multicolumn{1}{r|}{8}                 & 0.578                                                  \\
PSO {[}0, 1{]}                                         & \multicolumn{1}{r|}{4}               & \multicolumn{1}{r|}{2}                 & \textbf{\textless 0.001 *}                             & \multicolumn{1}{r|}{4.5}             & \multicolumn{1}{r|}{1}                 & 0.930                                                   & \multicolumn{1}{r|}{4}               & \multicolumn{1}{r|}{5}                 & \textbf{\textless 0.001 *}                             & \multicolumn{1}{r|}{5.5}             & \multicolumn{1}{r|}{5}                 & \textbf{\textless 0.001 *}                             \\
PSO - convex                                           & \multicolumn{1}{r|}{4}               & \multicolumn{1}{r|}{5}                 & \textbf{\textless 0.001 *}                             & \multicolumn{1}{r|}{6}               & \multicolumn{1}{r|}{2}                 & 0.837                                                  & \multicolumn{1}{r|}{1}               & \multicolumn{1}{r|}{4}                 & \textbf{\textless 0.001 *}                             & \multicolumn{1}{r|}{5.5}             & \multicolumn{1}{r|}{5}                 & \textbf{\textless 0.001 *}                             \\
Average                                                & \multicolumn{1}{r|}{9}               & \multicolumn{1}{r|}{8}                 & \textbf{0.01 *}                                        & \multicolumn{1}{r|}{4.5}             & \multicolumn{1}{r|}{3}                 & 0.703                                                  & \multicolumn{1}{r|}{8}               & \multicolumn{1}{r|}{8}                 & 0.660                                                   & \multicolumn{1}{r|}{9}               & \multicolumn{1}{r|}{9}                 & 0.120                                                   \\
RE                                                     & \multicolumn{1}{r|}{2}               & \multicolumn{1}{r|}{4}                 & \textbf{\textless 0.001 *}                             & \multicolumn{1}{r|}{3}               & \multicolumn{1}{r|}{4}                 & 0.464                                                  & \multicolumn{1}{r|}{5}               & \multicolumn{1}{r|}{1}                 & \textbf{\textless 0.001 *}                             & \multicolumn{1}{r|}{2.5}             & \multicolumn{1}{r|}{3}                 & \textbf{\textless 0.001 *}                            \\
\hline
\end{tabular}}

\caption{Performance rank of each method based on the location of the households and the respective p-value obtained from the Mann–Whitney U test. Very small p-values (less than 0.001) are shown as \textless 0.001. A p-value  that determines a method's forecast performance is significantly different across the two locations (i.e., p-value \textless 0.05) is marked in boldface with the *. }
\label{tab:locationwise_scores}
\end{sidewaystable}

\section{Conclusions}\label{sec:conclusion}

Forecasting the power generation of small-scale, distributed PV systems at different resolutions and horizons provides benefits to many stakeholders in the energy sector. In this study, we propose and evaluate a number of different forecasting approaches that improve PV power forecasting across multiple resolutions and horizons. We propose a forecast combination approach based on particle swarm optimization (PSO) that allows a forecaster to choose the best forecast combination regardless of the forecast resolution and horizon of interest.

We use five diverse base forecasters: seasonal naive, (seasonal) auto-regressive integrated moving average, (seasonal) auto-regressive integrated moving average with exogenous inputs, multiple linear regression, and support vector regression. However, the combination approach is independent of the type of the base forecaster and therefore any type of forecasting model can be incorporated as a base forecasting model. We explore several different conditions that can be imposed on the weights that are assigned to each of the base forecasters (such as allowing weights to be outside of the standard [0, 1] range), and explore how these conditions impact forecast performance. 

The forecast combination approaches are evaluated using real residential solar PV power data measured at 25 houses located in three different locations in the United States. The results show that the PSO-based forecast combination using weights in the range [0, 1] has the best overall forecast performance across the different resolutions and horizons studied. Furthermore, the PSO-based forecast combination approach can on average reduce the Mean Absolute Scaled Error by 3.81\% when compared to the best performing base forecaster. In future work it would be interesting to explore the forecast combination performance when the deterministic forecasts are extended to the context of probabilistic forecasts.

\section*{Acknowledgement}
This work was supported by the Melbourne Research Scholarship and an IBM Research Australia internship awarded to the first author and Australian Research Council Grant DP220101035 awarded to the last author. The research was undertaken using the LIEF HPC-GPGPU Facility hosted at the University of Melbourne. This Facility was established with the assistance of LIEF Grant LE170100200.

\bibliography{main_reference, references2}

\newpage
\appendix
\numberwithin{figure}{section}
\renewcommand*\thefigure{\Alph{section}.\arabic{figure}}

\section{Choosing the appropriate amount of recent data for high resolution ARIMA forecasts}\label{app:arima}

\begin{figure}[!th]
    \centering
    \includegraphics[width=0.9\textwidth]{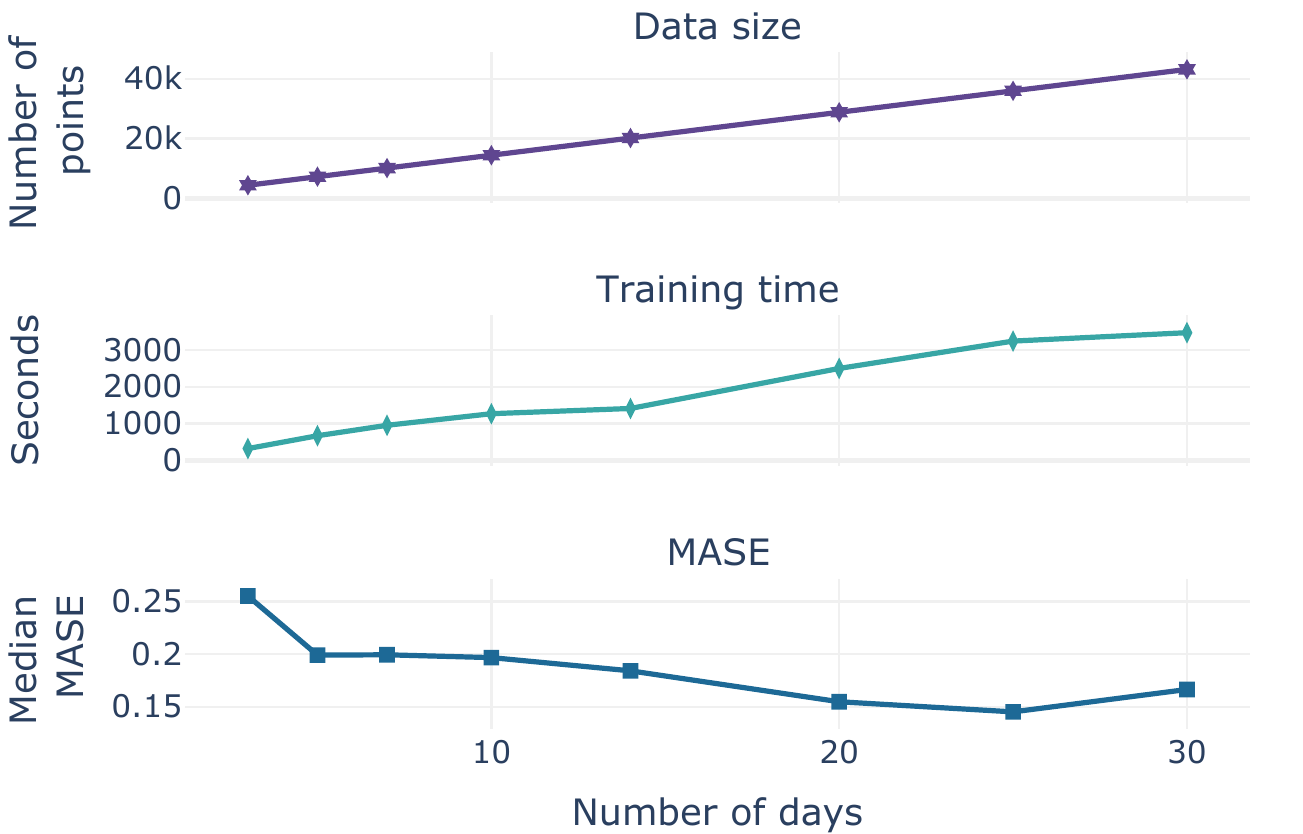}
    \caption{Data size of the samples, training time of the \textit{Auto Arima} function, and MASE to forecast a 5-minute horizon at a 1-minute resolution}
    \label{fig:higres-results}
\end{figure}

When dealing with high resolutions (1 minute and 5 minutes), two problems were encountered with regard to ARIMA models: (i) ~\textit{Auto ARIMA} runs out of memory when dealing with long seasonalities (e.g. for minutely data the seasonality would be 1440 due to the daily seasonal patterns)~\cite{hyndman2018forecasting}, (ii) training time significantly increases due to the large volume of data at these resolutions (e.g. for minutely data there are $100K+$ points)~\cite{wang2020distributed}. As a solution to problem (i), the seasonality of the data was added as a Fourier series to the \textit{Auto ARIMA} function, which is an effective way to significantly reduce memory requirements. To deal with problem (ii) an experimental analysis was conducted to verify the trade off between the training time and MASE for different data sizes. For this analysis, 25 independent samples across the time series of a household were considered. 
% Figure~\ref{fig:higres-results2} shows the distribution of the training time and the MASE across the samples to forecast 5 minute horizons at a 1-minute resolution. 
Figure~\ref{fig:higres-results} shows the number of data points for the respective training days and the median training time and MASE across the samples to forecast 5-minute horizons at a 1-minute resolution. The training time increases significantly with the increasing number of data points. The MASE decreases as the amount of historical training data is increased; however, beyond a certain value the reduction in the MASE becomes negligible.

As a result, for SARIMA models we use 25 days of recent data to forecast a 5-minute horizon with minutely data, and 14 days of recent data to forecast a 1-hour horizon at a 5-minute resolution.

\section{}\label{sec:appendixB}
Tables~\ref{tab:appendixTab1},~\ref{tab:appendixTab2},~\ref{tab:appendixTab3},~\ref{tab:appendixTab4} shows the mean MASE across all samples in the testing set (out-of-sample) of the 25 households studied in this paper.

\begin{table}[!h]
\centering
\caption{Mean MASE across all test samples (out-of-samples) for a 3-day horizon at a 1-day resolution}
\label{tab:appendixTab1}
\small
\begin{tabular}{|l|r|r|r|r|r|r|r|r|r|r|} 
\hline
\begin{tabular}[c]{@{}c@{}}\\House \\ID\end{tabular} & \multicolumn{1}{l|}{SN} & \multicolumn{1}{l|}{\begin{tabular}[c]{@{}l@{}}(s)\\ARIMA\end{tabular}} & \multicolumn{1}{l|}{\begin{tabular}[c]{@{}l@{}}(s)\\ARIMAX\end{tabular}} & \multicolumn{1}{l|}{MLR} & \multicolumn{1}{l|}{SVR} & \multicolumn{1}{l|}{PSO} & \multicolumn{1}{l|}{\begin{tabular}[c]{@{}l@{}}PSO\\~{[}0,1]\end{tabular}} & \multicolumn{1}{l|}{\begin{tabular}[c]{@{}l@{}}PSO- \\convex\end{tabular}} & \multicolumn{1}{l|}{Average} & \multicolumn{1}{l|}{RE}  \\ 
\hline
1                                                    & 1.561                   & 1.314                                                                   & 0.975                                                                    & 0.897                    & 0.992                    & 1.023                    & 1.008                                                                      & 1.027                                                                      & 1.104                        & 1.056                    \\
2                                                    & 1.616                   & 1.416                                                                   & 1.067                                                                    & 1.001                    & 1.177                    & 1.157                    & 1.006                                                                      & 1.032                                                                      & 1.217                        & 1.093                    \\
3                                                    & 1.508                   & 1.405                                                                   & 1.021                                                                    & 0.952                    & 1.081                    & 1.096                    & 1.049                                                                      & 1.068                                                                      & 1.148                        & 1.059                    \\
4                                                    & 0.895                   & 0.731                                                                   & 0.671                                                                    & 0.530                    & 0.806                    & 0.699                    & 0.589                                                                      & 0.631                                                                      & 0.643                        & 0.644                    \\
5                                                    & 1.177                   & 0.956                                                                   & 0.842                                                                    & 0.737                    & 0.947                    & 0.940                    & 0.781                                                                      & 0.842                                                                      & 0.877                        & 0.883                    \\
6                                                    & 1.113                   & 1.237                                                                   & 0.753                                                                    & 0.659                    & 0.805                    & 0.759                    & 0.730                                                                      & 0.761                                                                      & 0.827                        & 0.781                    \\
7                                                    & 1.450                   & 1.287                                                                   & 0.941                                                                    & 0.894                    & 1.019                    & 1.174                    & 0.958                                                                      & 0.980                                                                      & 1.081                        & 0.941                    \\
8                                                    & 2.157                   & 1.452                                                                   & 0.586                                                                    & 0.667                    & 0.545                    & 0.905                    & 0.586                                                                      & 0.586                                                                      & 0.978                        & 0.586                    \\
9                                                    & 1.700                   & 1.177                                                                   & 0.539                                                                    & 0.570                    & 0.513                    & 0.628                    & 0.539                                                                      & 0.539                                                                      & 0.796                        & 0.512                    \\
10                                                   & 1.400                   & 1.344                                                                   & 0.922                                                                    & 0.881                    & 0.961                    & 0.991                    & 0.901                                                                      & 0.922                                                                      & 1.046                        & 0.935                    \\
11                                                   & 1.992                   & 1.350                                                                   & 0.563                                                                    & 0.600                    & 0.548                    & 0.653                    & 0.563                                                                      & 0.563                                                                      & 0.915                        & 0.563                    \\
12                                                   & 1.581                   & 1.436                                                                   & 1.018                                                                    & 0.973                    & 1.113                    & 1.140                    & 1.032                                                                      & 1.062                                                                      & 1.177                        & 1.054                    \\
13                                                   & 1.477                   & 1.295                                                                   & 0.926                                                                    & 0.882                    & 1.011                    & 1.030                    & 0.920                                                                      & 0.948                                                                      & 1.068                        & 0.976                    \\
14                                                   & 1.711                   & 1.171                                                                   & 0.518                                                                    & 0.525                    & 0.442                    & 0.701                    & 0.518                                                                      & 0.518                                                                      & 0.776                        & 0.518                    \\
15                                                   & 1.511                   & 1.027                                                                   & 0.401                                                                    & 0.484                    & 0.413                    & 0.617                    & 0.506                                                                      & 0.461                                                                      & 0.632                        & 0.378                    \\
16                                                   & 1.184                   & 1.248                                                                   & 0.835                                                                    & 0.733                    & 0.909                    & 1.195                    & 0.898                                                                      & 0.963                                                                      & 0.909                        & 0.916                    \\
17                                                   & 1.482                   & 1.306                                                                   & 0.963                                                                    & 0.869                    & 1.053                    & 1.048                    & 1.049                                                                      & 1.093                                                                      & 1.093                        & 1.045                    \\
18                                                   & 1.163                   & 0.975                                                                   & 0.866                                                                    & 0.758                    & 0.935                    & 0.909                    & 0.807                                                                      & 0.838                                                                      & 0.886                        & 0.887                    \\
19                                                   & 1.349                   & 1.195                                                                   & 0.848                                                                    & 0.814                    & 0.903                    & 0.954                    & 0.913                                                                      & 0.944                                                                      & 0.975                        & 0.894                    \\
20                                                   & 1.520                   & 1.032                                                                   & 0.684                                                                    & 0.486                    & 0.507                    & 0.676                    & 0.690                                                                      & 0.683                                                                      & 0.649                        & 0.684                    \\
21                                                   & 2.130                   & 1.429                                                                   & 0.555                                                                    & 0.662                    & 0.552                    & 0.678                    & 0.555                                                                      & 0.555                                                                      & 0.974                        & 0.555                    \\
22                                                   & 2.158                   & 1.458                                                                   & 0.657                                                                    & 0.633                    & 0.564                    & 1.129                    & 0.960                                                                      & 0.593                                                                      & 0.951                        & 0.657                    \\
23                                                   & 1.839                   & 1.253                                                                   & 0.527                                                                    & 0.569                    & 0.479                    & 1.058                    & 0.527                                                                      & 0.527                                                                      & 0.837                        & 0.527                    \\
24                                                   & 1.321                   & 0.916                                                                   & 0.508                                                                    & 0.605                    & 0.501                    & 0.573                    & 0.485                                                                      & 0.508                                                                      & 0.609                        & 0.451                    \\
25                                                   & 2.911                   & 2.354                                                                   & 1.054                                                                    & 1.898                    & 1.932                    & 2.981                    & 1.275                                                                      & 1.916                                                                      & 1.766                        & 1.333                    \\
\hline
\end{tabular}
\end{table}

\begin{table}[!h]
\centering
\caption{Mean MASE across all test samples (out-of-samples) for a 1-day horizon at a 1-hour resolution}
\label{tab:appendixTab2}
\small
\begin{tabular}{|l|r|r|r|r|r|r|r|r|r|r|} 
\hline
\begin{tabular}[c]{@{}c@{}}\\House\\ID\end{tabular} & \multicolumn{1}{l|}{SN} & \multicolumn{1}{l|}{\begin{tabular}[c]{@{}l@{}}(s)\\ARIMA\end{tabular}} & \multicolumn{1}{l|}{\begin{tabular}[c]{@{}l@{}}(s)\\ARIMAX\end{tabular}} & \multicolumn{1}{l|}{MLR} & \multicolumn{1}{l|}{SVR} & \multicolumn{1}{l|}{PSO} & \multicolumn{1}{l|}{\begin{tabular}[c]{@{}l@{}}PSO\\{[}0,1]\end{tabular}} & \multicolumn{1}{l|}{\begin{tabular}[c]{@{}l@{}}PSO-\\convex\end{tabular}} & \multicolumn{1}{l|}{Average} & \multicolumn{1}{l|}{RE}  \\ 
\hline
1                                                   & 1.208                   & 1.229                                                                   & 1.150                                                                    & 1.398                    & 1.381                    & 1.379                    & 1.101                                                                     & 1.176                                                                     & 1.189                        & 1.135                    \\
2                                                   & 1.339                   & 1.268                                                                   & 1.199                                                                    & 1.524                    & 1.472                    & 1.316                    & 1.329                                                                     & 1.218                                                                     & 1.283                        & 1.212                    \\
3                                                   & 1.281                   & 1.227                                                                   & 1.161                                                                    & 1.481                    & 1.403                    & 1.214                    & 1.212                                                                     & 1.166                                                                     & 1.231                        & 1.158                    \\
4                                                   & 0.709                   & 1.224                                                                   & 1.130                                                                    & 0.747                    & 1.132                    & 0.728                    & 0.855                                                                     & 0.898                                                                     & 0.712                        & 0.870                    \\
5                                                   & 0.970                   & 1.205                                                                   & 1.048                                                                    & 0.984                    & 1.269                    & 0.990                    & 0.918                                                                     & 0.942                                                                     & 0.928                        & 0.997                    \\
6                                                   & 0.847                   & 1.253                                                                   & 0.993                                                                    & 1.026                    & 1.140                    & 1.067                    & 0.841                                                                     & 0.896                                                                     & 0.856                        & 0.928                    \\
7                                                   & 1.128                   & 1.294                                                                   & 1.154                                                                    & 1.473                    & 1.397                    & 1.251                    & 1.076                                                                     & 1.076                                                                     & 1.139                        & 1.106                    \\
8                                                   & 1.270                   & 1.666                                                                   & 1.392                                                                    & 1.474                    & 0.953                    & 1.295                    & 1.289                                                                     & 1.289                                                                     & 1.182                        & 1.186                    \\
9                                                   & 1.048                   & 1.708                                                                   & 1.271                                                                    & 1.131                    & 0.904                    & 0.941                    & 1.008                                                                     & 1.021                                                                     & 1.012                        & 1.006                    \\
10                                                  & 1.198                   & 1.314                                                                   & 1.214                                                                    & 1.619                    & 1.380                    & 1.295                    & 1.205                                                                     & 1.184                                                                     & 1.252                        & 1.152                    \\
11                                                  & 1.177                   & 1.703                                                                   & 1.374                                                                    & 1.344                    & 0.921                    & 1.265                    & 1.221                                                                     & 1.209                                                                     & 1.122                        & 1.132                    \\
12                                                  & 1.300                   & 1.416                                                                   & 1.274                                                                    & 1.598                    & 1.348                    & 1.257                    & 1.234                                                                     & 1.234                                                                     & 1.211                        & 1.251                    \\
13                                                  & 1.244                   & 1.376                                                                   & 1.201                                                                    & 1.417                    & 1.267                    & 1.525                    & 1.252                                                                     & 1.252                                                                     & 1.109                        & 1.190                    \\
14                                                  & 1.028                   & 1.878                                                                   & 1.313                                                                    & 1.165                    & 0.890                    & 0.984                    & 1.016                                                                     & 1.016                                                                     & 1.044                        & 0.992                    \\
15                                                  & 0.953                   & 1.813                                                                   & 1.135                                                                    & 0.847                    & 0.961                    & 0.897                    & 0.861                                                                     & 0.861                                                                     & 0.910                        & 0.840                    \\
16                                                  & 0.993                   & 1.333                                                                   & 1.055                                                                    & 0.998                    & 1.094                    & 1.000                    & 1.041                                                                     & 1.066                                                                     & 0.896                        & 1.004                    \\
17                                                  & 1.184                   & 1.316                                                                   & 1.094                                                                    & 1.340                    & 1.392                    & 1.214                    & 1.106                                                                     & 1.105                                                                     & 1.102                        & 1.108                    \\
18                                                  & 0.973                   & 1.405                                                                   & 1.316                                                                    & 1.153                    & 1.326                    & 1.445                    & 1.077                                                                     & 1.077                                                                     & 0.941                        & 1.089                    \\
19                                                  & 1.123                   & 1.252                                                                   & 1.119                                                                    & 1.368                    & 1.164                    & 1.097                    & 1.073                                                                     & 1.073                                                                     & 1.064                        & 1.074                    \\
20                                                  & 0.949                   & 1.763                                                                   & 1.237                                                                    & 1.035                    & 0.975                    & 0.898                    & 0.891                                                                     & 0.890                                                                     & 0.972                        & 0.896                    \\
21                                                  & 1.302                   & 1.759                                                                   & 1.342                                                                    & 1.379                    & 0.938                    & 1.266                    & 1.281                                                                     & 1.270                                                                     & 1.155                        & 1.200                    \\
22                                                  & 1.260                   & 1.709                                                                   & 1.573                                                                    & 1.605                    & 1.004                    & 1.381                    & 1.393                                                                     & 1.373                                                                     & 1.268                        & 1.281                    \\
23                                                  & 1.110                   & 1.671                                                                   & 1.377                                                                    & 1.238                    & 0.851                    & 1.279                    & 1.086                                                                     & 1.087                                                                     & 1.058                        & 1.078                    \\
24                                                  & 0.819                   & 1.502                                                                   & 1.404                                                                    & 0.961                    & 0.715                    & 0.849                    & 0.836                                                                     & 0.843                                                                     & 0.846                        & 0.768                    \\
25                                                  & 1.705                   & 4.356                                                                   & 3.615                                                                    & 3.529                    & 2.253                    & 2.676                    & 1.755                                                                     & 1.755                                                                     & 2.845                        & 1.860                    \\
\hline
\end{tabular}
\end{table}

\begin{table}
\centering
\small
\caption{Mean MASE across all test samples (out-of-samples) for a 1-hour horizon at a 5-minute resolution}
\label{tab:appendixTab3}
\begin{tabular}{|l|r|r|r|r|r|r|r|r|r|r|} 
\hline
\begin{tabular}[c]{@{}c@{}}\\House \\ID\end{tabular} & \multicolumn{1}{l|}{SN} & \multicolumn{1}{l|}{\begin{tabular}[c]{@{}l@{}}(s)\\ARIMA\end{tabular}} & \multicolumn{1}{l|}{\begin{tabular}[c]{@{}l@{}}(s)\\ARIMAX\end{tabular}} & \multicolumn{1}{l|}{MLR} & \multicolumn{1}{l|}{SVR} & \multicolumn{1}{l|}{PSO} & \multicolumn{1}{l|}{\begin{tabular}[c]{@{}l@{}}PSO\\~{[}0,1]\end{tabular}} & \multicolumn{1}{l|}{\begin{tabular}[c]{@{}l@{}}PSO- \\convex\end{tabular}} & \multicolumn{1}{l|}{Average} & \multicolumn{1}{l|}{RE}  \\ 
\hline
1                                                    & 1.036                   & 0.625                                                                   & 0.625                                                                    & 1.821                    & 2.300                    & 0.893                    & 0.625                                                                      & 0.625                                                                      & 1.004                        & 0.621                    \\
2                                                    & 1.080                   & 0.654                                                                   & 0.652                                                                    & 1.977                    & 2.607                    & 0.710                    & 0.654                                                                      & 0.654                                                                      & 1.106                        & 0.654                    \\
3                                                    & 1.079                   & 0.615                                                                   & 0.616                                                                    & 1.928                    & 2.330                    & 0.892                    & 0.615                                                                      & 0.615                                                                      & 1.045                        & 0.615                    \\
4                                                    & 0.787                   & 0.443                                                                   & 0.437                                                                    & 1.047                    & 1.752                    & 1.074                    & 0.443                                                                      & 0.443                                                                      & 0.678                        & 0.432                    \\
5                                                    & 0.904                   & 0.526                                                                   & 0.524                                                                    & 1.457                    & 2.047                    & 0.682                    & 0.524                                                                      & 0.524                                                                      & 0.836                        & 0.526                    \\
6                                                    & 0.867                   & 0.499                                                                   & 0.497                                                                    & 1.316                    & 1.941                    & 0.726                    & 0.499                                                                      & 0.499                                                                      & 0.788                        & 0.497                    \\
7                                                    & 1.030                   & 0.622                                                                   & 0.620                                                                    & 1.810                    & 2.243                    & 0.670                    & 0.620                                                                      & 0.620                                                                      & 0.986                        & 0.613                    \\
8                                                    & 1.451                   & 0.728                                                                   & 0.771                                                                    & 2.127                    & 1.163                    & 0.826                    & 0.728                                                                      & 0.728                                                                      & 1.001                        & 0.727                    \\
9                                                    & 1.394                   & 0.720                                                                   & 0.732                                                                    & 1.649                    & 1.122                    & 0.847                    & 0.738                                                                      & 0.710                                                                      & 0.918                        & 0.721                    \\
10                                                   & 0.918                   & 0.573                                                                   & 0.579                                                                    & 1.786                    & 2.299                    & 1.417                    & 0.573                                                                      & 0.573                                                                      & 0.998                        & 0.564                    \\
11                                                   & 1.458                   & 0.697                                                                   & 0.729                                                                    & 1.941                    & 1.148                    & 0.786                    & 0.697                                                                      & 0.697                                                                      & 0.974                        & 0.698                    \\
12                                                   & 1.153                   & 0.704                                                                   & 0.694                                                                    & 2.091                    & 2.546                    & 0.710                    & 0.698                                                                      & 0.698                                                                      & 1.190                        & 0.697                    \\
13                                                   & 1.024                   & 0.612                                                                   & 0.637                                                                    & 1.805                    & 2.243                    & 0.896                    & 0.605                                                                      & 0.605                                                                      & 1.050                        & 0.601                    \\
14                                                   & 1.428                   & 0.676                                                                   & 0.690                                                                    & 1.590                    & 1.075                    & 0.808                    & 0.676                                                                      & 0.676                                                                      & 0.895                        & 0.676                    \\
15                                                   & 1.100                   & 0.651                                                                   & 0.666                                                                    & 1.245                    & 1.384                    & 0.849                    & 0.717                                                                      & 0.663                                                                      & 0.852                        & 0.651                    \\
16                                                   & 0.935                   & 0.544                                                                   & 0.542                                                                    & 1.515                    & 2.103                    & 0.767                    & 0.633                                                                      & 0.630                                                                      & 0.878                        & 0.544                    \\
17                                                   & 1.065                   & 0.559                                                                   & 0.556                                                                    & 1.764                    & 1.970                    & 1.158                    & 0.557                                                                      & 0.557                                                                      & 0.963                        & 0.557                    \\
18                                                   & 0.966                   & 0.568                                                                   & 0.563                                                                    & 1.356                    & 2.267                    & 0.794                    & 0.648                                                                      & 0.647                                                                      & 0.852                        & 0.554                    \\
19                                                   & 0.881                   & 0.571                                                                   & 0.578                                                                    & 1.787                    & 2.140                    & 1.038                    & 0.624                                                                      & 0.599                                                                      & 1.003                        & 0.571                    \\
20                                                   & 1.483                   & 0.757                                                                   & 0.795                                                                    & 1.718                    & 1.301                    & 1.677                    & 0.763                                                                      & 0.763                                                                      & 0.915                        & 0.751                    \\
21                                                   & 1.475                   & 0.739                                                                   & 0.694                                                                    & 2.027                    & 1.272                    & 1.170                    & 0.739                                                                      & 0.739                                                                      & 0.995                        & 0.731                    \\
22                                                   & 1.642                   & 0.813                                                                   & 0.875                                                                    & 2.375                    & 1.406                    & 0.820                    & 0.813                                                                      & 0.813                                                                      & 1.145                        & 0.807                    \\
23                                                   & 1.354                   & 0.656                                                                   & 0.670                                                                    & 1.685                    & 1.052                    & 0.690                    & 0.636                                                                      & 0.636                                                                      & 0.874                        & 0.644                    \\
24                                                   & 1.565                   & 0.710                                                                   & 0.687                                                                    & 1.646                    & 1.335                    & 0.740                    & 0.710                                                                      & 0.710                                                                      & 0.932                        & 0.737                    \\
25                                                   & 3.219                   & 1.484                                                                   & 1.856                                                                    & 4.945                    & 3.047                    & 2.305                    & 1.508                                                                      & 1.655                                                                      & 2.048                        & 1.450                    \\
\hline
\end{tabular}
\end{table}

\begin{table}
\centering
\small
\caption{Mean MASE across all test samples (out-of-samples) for a 5-minute horizon at a 1-minute resolution}
\label{tab:appendixTab4}
\begin{tabular}{|l|r|r|r|r|r|r|r|r|r|r|} 
\hline
\begin{tabular}[c]{@{}c@{}}\\House\\~ID\end{tabular} & \multicolumn{1}{l|}{SN} & \multicolumn{1}{l|}{\begin{tabular}[c]{@{}l@{}}(s)\\ARIMA\end{tabular}} & \multicolumn{1}{l|}{\begin{tabular}[c]{@{}l@{}}(s)\\ARIMAX\end{tabular}} & \multicolumn{1}{l|}{MLR} & \multicolumn{1}{l|}{SVR} & \multicolumn{1}{l|}{PSO} & \multicolumn{1}{l|}{\begin{tabular}[c]{@{}l@{}}PSO \\{[}0,1]\end{tabular}} & \multicolumn{1}{l|}{\begin{tabular}[c]{@{}l@{}}PSO- \\convex\end{tabular}} & \multicolumn{1}{l|}{Average} & \multicolumn{1}{l|}{RE}  \\ 
\hline
1                                                    & 0.259                   & 0.251                                                                   & 0.257                                                                    & 1.914                    & 3.529                    & 1.023                    & 0.251                                                                      & 0.251                                                                      & 0.883                        & 0.251                    \\
2                                                    & 0.273                   & 0.265                                                                   & 0.270                                                                    & 2.072                    & 3.631                    & 0.900                    & 0.265                                                                      & 0.265                                                                      & 0.960                        & 0.255                    \\
3                                                    & 0.283                   & 0.267                                                                   & 0.274                                                                    & 2.045                    & 3.148                    & 0.980                    & 0.267                                                                      & 0.267                                                                      & 0.833                        & 0.267                    \\
4                                                    & 0.186                   & 0.169                                                                   & 0.164                                                                    & 0.986                    & 2.448                    & 0.623                    & 0.169                                                                      & 0.169                                                                      & 0.597                        & 0.169                    \\
5                                                    & 0.207                   & 0.192                                                                   & 0.194                                                                    & 1.469                    & 3.278                    & 0.725                    & 0.192                                                                      & 0.192                                                                      & 0.757                        & 0.192                    \\
6                                                    & 0.234                   & 0.246                                                                   & 0.246                                                                    & 1.337                    & 2.811                    & 0.572                    & 0.246                                                                      & 0.246                                                                      & 0.682                        & 0.246                    \\
7                                                    & 0.220                   & 0.167                                                                   & 0.171                                                                    & 1.884                    & 2.940                    & 0.235                    & 0.167                                                                      & 0.167                                                                      & 0.832                        & 0.167                    \\
8                                                    & 0.415                   & 0.298                                                                   & 0.297                                                                    & 2.252                    & 1.525                    & 0.741                    & 0.387                                                                      & 0.374                                                                      & 0.846                        & 0.298                    \\
9                                                    & 0.403                   & 0.302                                                                   & 0.299                                                                    & 1.634                    & 1.220                    & 0.592                    & 0.318                                                                      & 0.318                                                                      & 0.672                        & 0.302                    \\
10                                                   & 0.227                   & 0.239                                                                   & 0.248                                                                    & 1.868                    & 3.309                    & 0.961                    & 0.239                                                                      & 0.239                                                                      & 0.976                        & 0.239                    \\
11                                                   & 0.404                   & 0.296                                                                   & 0.293                                                                    & 1.991                    & 1.467                    & 0.684                    & 0.296                                                                      & 0.296                                                                      & 0.767                        & 0.296                    \\
12                                                   & 0.275                   & 0.268                                                                   & 0.269                                                                    & 2.189                    & 3.323                    & 1.056                    & 0.269                                                                      & 0.269                                                                      & 0.960                        & 0.268                    \\
13                                                   & 0.219                   & 0.228                                                                   & 0.226                                                                    & 1.896                    & 3.134                    & 0.599                    & 0.226                                                                      & 0.226                                                                      & 0.860                        & 0.226                    \\
14                                                   & 0.425                   & 0.302                                                                   & 0.301                                                                    & 1.586                    & 1.294                    & 0.652                    & 0.302                                                                      & 0.302                                                                      & 0.656                        & 0.302                    \\
15                                                   & 0.430                   & 0.327                                                                   & 0.348                                                                    & 1.357                    & 1.796                    & 1.175                    & 0.342                                                                      & 0.342                                                                      & 0.793                        & 0.337                    \\
16                                                   & 0.187                   & 0.156                                                                   & 0.158                                                                    & 1.581                    & 3.115                    & 0.448                    & 0.158                                                                      & 0.158                                                                      & 0.752                        & 0.158                    \\
17                                                   & 0.263                   & 0.252                                                                   & 0.251                                                                    & 1.973                    & 2.530                    & 0.852                    & 0.252                                                                      & 0.252                                                                      & 0.779                        & 0.252                    \\
18                                                   & 0.226                   & 0.253                                                                   & 0.257                                                                    & 1.460                    & 3.174                    & 0.578                    & 0.255                                                                      & 0.255                                                                      & 0.878                        & 0.252                    \\
19                                                   & 0.184                   & 0.179                                                                   & 0.180                                                                    & 2.007                    & 3.167                    & 1.081                    & 0.179                                                                      & 0.179                                                                      & 0.833                        & 0.179                    \\
20                                                   & 0.357                   & 0.270                                                                   & 0.271                                                                    & 1.645                    & 1.315                    & 0.551                    & 0.289                                                                      & 0.289                                                                      & 0.562                        & 0.276                    \\
21                                                   & 0.427                   & 0.314                                                                   & 0.305                                                                    & 2.094                    & 1.492                    & 0.570                    & 0.358                                                                      & 0.358                                                                      & 0.833                        & 0.314                    \\
22                                                   & 0.459                   & 0.351                                                                   & 0.346                                                                    & 2.567                    & 1.805                    & 0.847                    & 0.415                                                                      & 0.436                                                                      & 0.984                        & 0.356                    \\
23                                                   & 0.376                   & 0.266                                                                   & 0.262                                                                    & 1.708                    & 1.363                    & 0.575                    & 0.286                                                                      & 0.286                                                                      & 0.685                        & 0.266                    \\
24                                                   & 0.429                   & 0.329                                                                   & 0.326                                                                    & 1.588                    & 1.541                    & 0.932                    & 0.329                                                                      & 0.329                                                                      & 0.670                        & 0.325                    \\
25                                                   & 0.872                   & 0.666                                                                   & 0.658                                                                    & 4.974                    & 3.141                    & 2.330                    & 0.666                                                                      & 0.666                                                                      & 1.715                        & 0.666                    \\
\hline
\end{tabular}
\end{table}

\end{document}